\documentclass{article}
\usepackage{graphicx}
\usepackage{amssymb} 
\usepackage{amsmath}  
\usepackage{algorithm}
\usepackage{algorithmic}
\usepackage{booktabs}
\usepackage{enumitem}
\usepackage{graphicx}
\usepackage{wrapfig}
\usepackage{multirow}
\usepackage{afterpage}
\usepackage{caption}
\usepackage{tabularx}
\usepackage{subcaption}
\usepackage{hyperref}
\usepackage{booktabs}
\usepackage[final]{corl_2026} 

\newcommand{\pio}{$\pi_0$}
\newcommand{\pihalf}{$\pi_{0.5}$}
\title{Barrier-Constrained Guidance For Real-time Asynchronous VLA Execution}
\author{
  \textbf{Wenbo Yu}\textsuperscript{1,3$*$}, \quad
  \textbf{Jifan Li}\textsuperscript{2,3$*$}, \quad
  \textbf{Wenke Xia}\textsuperscript{2,3$*$}, \quad
  \textbf{Yuchen Li}\textsuperscript{2}, \quad
  \textbf{Yizhuo Zhang}\textsuperscript{2,3}, \\
  \textbf{Pei Ren}\textsuperscript{3}, \quad
  \textbf{Zhengping Che}\textsuperscript{3,$\dagger$}, \quad
  \textbf{Di Hu}\textsuperscript{2,$\dagger$} 
  \vspace{0.2cm} \\ 
  \textsuperscript{1}Beijing Forestry University\\
  \textsuperscript{2}Gaoling School of Artificial Intelligence, Renmin University of China \\
  \textsuperscript{3}Beijing Innovation Center of Humanoid Robotics \\
  \vspace{-0.5cm} 
  \small{\textsuperscript{$*$}Equal contribution \quad 
  \textsuperscript{$\dagger$}Corresponding authors}
}

\begin{document}
\maketitle
\renewcommand{\thefootnote}{\fnsymbol{footnote}}
\footnotetext[2]{Corresponding authors:
\texttt{z.che@x-humanoid.com},
\texttt{dihu@ruc.edu.cn}}
\begin{abstract}
While asynchronous execution enables VLA models to operate in real-time environments by predicting the next chunk during execution, it inevitably introduces boundary misalignment with the committed action stream. Existing delay-aware solutions address this at the prohibitive cost of extensive retraining, architectural modifications, or fixed-prefix inpainting that inflates computation when low latency is vital. To address these problems, we propose \textit{FreeBar}, a \textbf{training-free real-time inference} plugin that enables robust asynchronous execution with only marginal additional inference cost by casting chunk alignment in flow-based VLAs as barrier-guided action-space constraint. Specifically, FreeBar introduces a lightweight Control Barrier Function (CBF)-inspired consistency constraint between consecutive action chunks and solves the resulting correction in closed form, avoiding costly backpropagation through the VLA action head. By progressively tightening the consistency bound during generation, FreeBar aligns generated chunks with the committed execution stream while preserving early-stage generative and within-chunk future action flexibility,. Across Kinetix, RoboCasa, and diverse real-world robotic platforms with multiple VLA backbones, FreeBar improves success rates over naive asynchronous execution by \textbf{10\%--15\%} on average, with larger gains in highly dynamic scenarios. It also preserves real-time computational efficiency, decreasing inference latency per action chunk over \textbf{45\%} than RTC. Additionally, as a universal plugin, FreeBar remains robust across distinct latency environments, base policies, and hyperparameter settings.
\end{abstract}

\keywords{Robot manipulation, Real-time inference} 

\section{Introduction}
Modern Vision-Language-Action (VLA) models ~\cite{gr00tn1_2025,black2026pi0visionlanguageactionflowmodel, intelligence2025pi05visionlanguageactionmodelopenworld, kim24openvla, shukor2025smolvla, zitkovich2023rt,fan2025xr,liu2026last0latentspatiotemporalchainofthought} have demonstrated remarkable capabilities in complex robotic manipulation. Although action chunking amortizes policy queries and captures short-horizon action correlations for smoother control~\cite{li2026reinforcementlearningactionchunking}, synchronous execution blocks robot control until a subsequent chunk is fully generated, thereby translating unavoidable inference latency into significant control pauses. Consequently, these intermittent pauses not only degrade motion quality by causing jerky trajectories but also induce critical state-action mismatches that frequently lead to irreversible task failures.

Asynchronous inference~\cite{ma2025runningvlasrealtimespeed,sendai2025leaveobservationbehindrealtime,black2026real, zou2025asynchronous} avoids this issue by overlapping policy inference with action execution, thereby enabling continuous control. However, asynchronous execution introduces a temporal misalignment between the state used for inference and the action sequence executed. Existing approaches to asynchronous VLA execution address this problem through either delay-aware policy training or online trajectory correction. Training-based methods typically require changes to the original VLA pipeline, such as delay-aware data construction~\cite{tang2025vlash}, auxiliary networks~\cite{sendai2025leaveobservationbehindrealtime}, or modified policy architectures~\cite{xie2026dynamicvla}, thereby limiting their scalability and general applicability across VLA architectures. Training-free methods, such as Real-Time Chunking (RTC)~\cite{black2026real}, avoid retraining by enforcing temporal consistency through fixed-prefix inpainting and vector-Jacobian-product guidance. While effective, this turns chunk alignment into an additional gradient-based correction process during inference, increasing computation when low latency is most critical.

To address these challenges, we propose FreeBar, a training-free barrier-constrained real-time inference framework for asynchronous VLA execution. While prior work \cite{black2026real} demonstrates that aligning the initial steps of a newly generated action chunk with currently executing actions effectively improves trajectory continuity, achieving this alignment without computationally expensive gradient evaluations remains an open challenge. Inspired by Control Barrier Functions (CBFs) of minimally modifying a nominal dynamics to satisfy a constraint as well as recent barrier-based regularization in robotic policy learning~\cite{sun2025latentpolicybarrierlearning}, we introduce a CBF-inspired temporal-consistency constraint over action chunks during generation. At the beginning of generation, a broad barrier is applied to preserve the underlying VLA's generative flexibility. As denoising progresses, this barrier gradually tightens toward the currently executing actions, seamlessly guiding the generated trajectory into temporal alignment. By constructing a lightweight correction directly from action-space discrepancies, rather than obtaining guidance through computationally expensive Vector-Jacobian Product (VJP) evaluations~\cite{black2026real}, FreeBar steers generated actions toward a temporally consistent feasible region during denoising to maintain action controllability. Consequently, this design incurs minimal overhead from the guidance computation and serves as a plug-and-play module for flow-based VLAs without necessitating any modifications to the underlying architecture or training pipeline.

\begin{figure}[t]
    \centering
    \includegraphics[width=\textwidth]{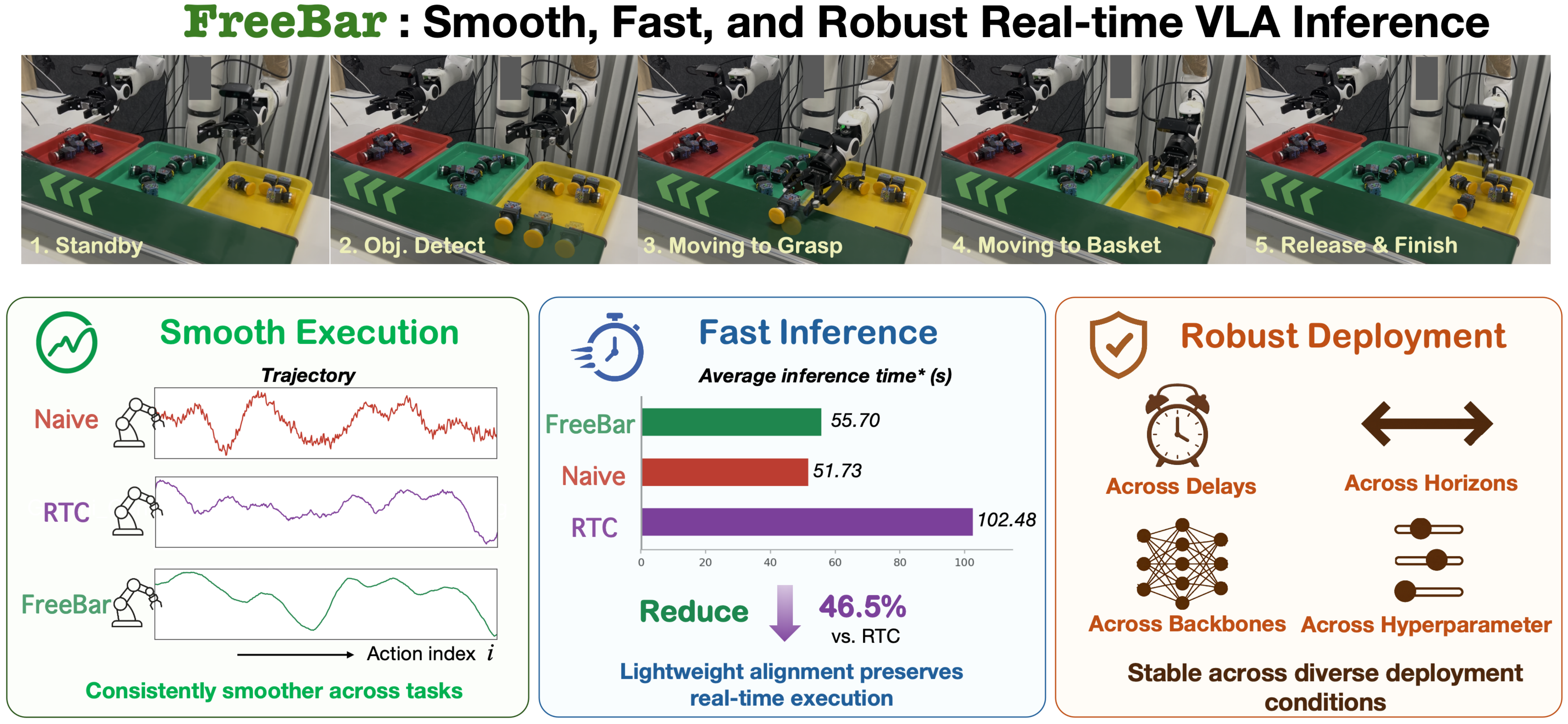}
    \captionsetup{skip=5pt}
    \caption{
        \textbf{FreeBar enables smooth, efficient, and robust asynchronous VLA execution.}
        \textbf{(Top)} Swift and fluent robotic responses to dynamic targets (e.g., a moving button).
        \textbf{(Bottom)} FreeBar ensures \textbf{smoother} kinematics than RTC, \textbf{efficient} inference via lightweight constraint over costly VJP, and \textbf{robust} stability across diverse latencies, frameworks, and hyperparameters.  ($^*$GR00T on RoboCasa, 4 denoising steps.)
    }
        \vspace{0.5cm}
    \label{fig:teaser}
\end{figure}

We evaluate FreeBar on high-dynamic tasks in Kinetix, RoboCasa, and real-world robotic platforms with multiple widely adopted VLAs including GR00T~\cite{gr00tn1_2025}, \pio~\cite{black2026pi0visionlanguageactionflowmodel} and \pihalf~\cite{intelligence2025pi05visionlanguageactionmodelopenworld}. 
FreeBar improves success rates over naive asynchronous execution by 10–15\% on average, while reducing inference time by over 45\% compared to RTC and remaining highly robust across varying latency environments, VLA backbones, and hyperparameter settings. These results indicate that closed-form barrier-constrained chunk alignment is an effective and robust training-free mechanism for asynchronous VLA deployment, improving temporal consistency with substantially lower inference cost than VJP-based correction.

\section{Related Work}

\subsection{Vision-Language-Action Models and Action Chunking}

Vision-Language-Action (VLA) models~\cite{gr00tn1_2025,black2026pi0visionlanguageactionflowmodel, intelligence2025pi05visionlanguageactionmodelopenworld, kim24openvla, shukor2025smolvla, team2025gemini} have become a central paradigm for generalist robotic manipulation, mapping visual observations and language instructions to executable robot actions~\cite{kawaharazuka2025vision}. Recent robot foundation models combine large-scale vision-language pretraining with robot demonstration data to improve generalization across tasks, objects, and environments. Many VLA policies further predict temporally extended action chunks rather than single-step actions~\cite{li2026reinforcementlearningactionchunking, zhao2023learning}, improving local smoothness, amortizing policy queries, and providing stronger temporal priors. However, action chunking complicates real-time deployment: since an entire chunk is generated from a single observation, inference latency can cause the predicted actions to be executed under a physical state that has already changed. This issue is especially pronounced for diffusion-based~\cite{hoeg2024streaming, chi2025diffusion, pearce2023imitating, frans2025one, liang2025discrete} and flow-based~\cite{black2026pi0visionlanguageactionflowmodel, chen2026dfm, jiang2025streaming,su2026freqpolicy} VLA policies, whose iterative generation introduces non-negligible delay. Real-time VLA deployment therefore requires not only accurate action prediction, but also temporal alignment between generated chunks and ongoing execution.

\subsection{Real-time Inference for VLA Models}

Asynchronous inference addresses real-time VLA deployment by overlapping policy inference with action execution~\cite{li2026reinforcementlearningactionchunking,ma2025runningvlasrealtimespeed,sendai2025leaveobservationbehindrealtime,black2026real, zou2025asynchronous}, avoiding the stop-and-go behavior of synchronous execution. However, it introduces a mismatch between the observation used for generation and the state where the resulting chunk is executed. Existing methods address this mismatch through either training-based latency adaptation~\cite{zhu2026deflect} or training-free inference-time correction. Training-based approaches use delayed data, future-state prediction~\cite{tang2025vlash, shi2026streamingvla}, auxiliary compensation modules~\cite{sendai2025leaveobservationbehindrealtime, zhu2026deflect, niu2026realtime}, or modified objectives~\cite{xie2026dynamicvla, yang2026dysl, jiang2025asyncvla}, but require access to the training pipeline and are difficult to apply to frozen VLA policies. Training-free methods instead modify generated actions during inference; for example, Real-Time Chunking (RTC)~\cite{black2026real} uses VJP-based guidance~\cite{belkhale2024minivla, lee2024behavior} to align consecutive chunks, but repeated differentiation increases inference overhead and introduces sensitivity to guidance strength and scheduling. FreeBar differs by replacing VJP-dependent correction with lightweight barrier-guided chunk alignment, introducing progressive consistency bounds during generation to preserve short-horizon continuity while leaving future actions adaptive.

\section{Method}

We propose FreeBar, a training-free inference plugin for real-time asynchronous VLA execution. 
Instead of using VJP-based gradient guidance, FreeBar applies a closed-form action-space projection that aligns the generated chunk with the ongoing execution stream. 
The projection imposes a feasible correction set that contracts over denoising time and across the prediction horizon, preserving short-term continuity while ensuring the generation fidelity of the entire action chunk.

\subsection{Problem Setup: Asynchronous VLA Inference}

We consider a generative VLA policy that predicts an action chunk :
\[
A=\{a_0,\dots,a_{H-1}\}\in\mathbb{R}^{H\times D}
\]
conditioned on multimodal context $c$.
For flow-based action generators, iterative chunk generation can be expressed as 
$d x_t / dt = v_\theta(x_t,t,c)$, where $x_t$ denotes the intermediate action chunk and $v_\theta$ is the velocity predicted by the VLA action head. 
Diffusion-based policies can be handled in the same form by interpreting $v_\theta$ as the sampler update direction. Pure autoregressive or one-step policies that do not expose such iterative generation dynamics are outside the current scope.

In asynchronous execution, the robot executes the current chunk while the policy generates the subsequent one. 
If inference takes $d$ control steps, the new chunk is conditioned on a stale observation and may be discontinuous with the actions already committed to the low-level controller. 
Let $\bar{A}=\{\bar{a}_0,\dots,\bar{a}_{H-1}\}$ denotes the time-aligned carry-over reference from the previous inference cycle. 
Real-time alignment should keep the proximal part of the new chunk consistent with $\bar{A}$ while avoiding over-constraining remaining actions. 
FreeBar addresses this trade-off using closed-form projection with generation-time tightening and horizon-wise graded constraints.

\begin{figure}[t] 
    \centering
    \includegraphics[width=1\textwidth]{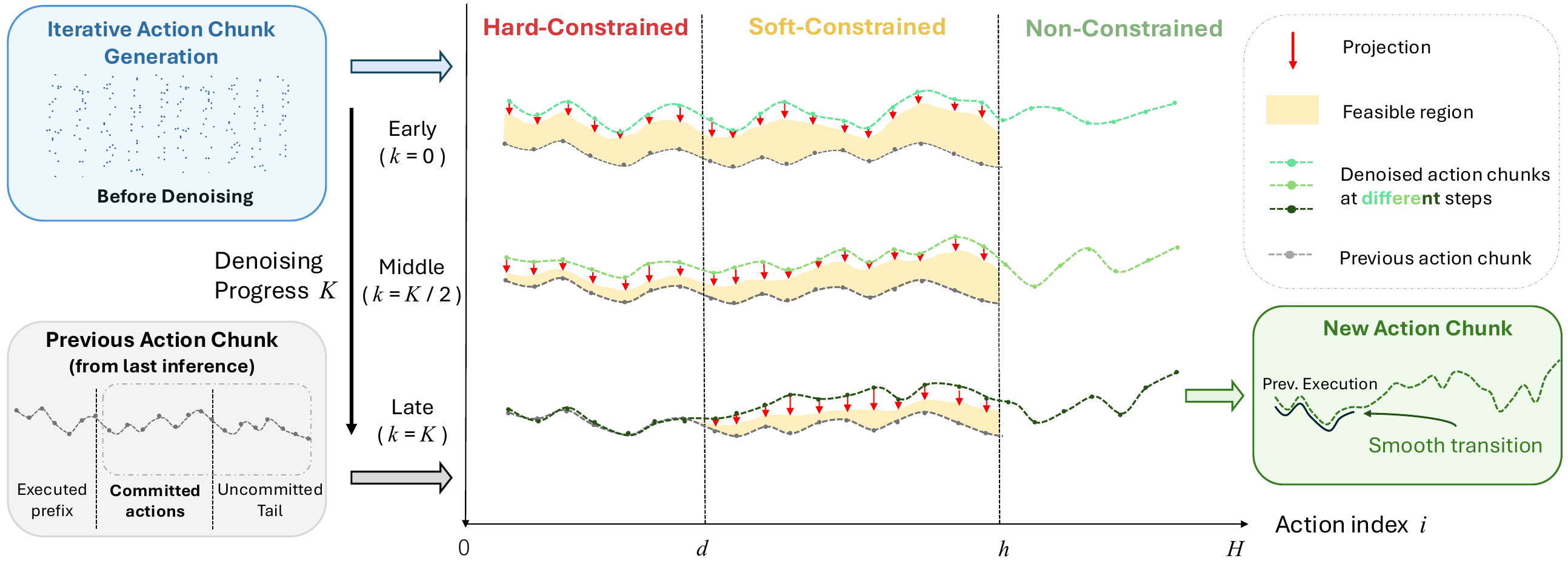} 
    \caption{\textbf{Progressive barrier-guided chunk alignment in FreeBar.}
    During asynchronous execution, FreeBar aligns the newly generated action chunk with a time-aligned reference from the previous chunk using closed-form action-space projection. The barrier constraint is progressively tightened across denoising steps and graded along the horizon: proximal actions are strongly aligned, medial actions are softly regularized, and distal actions remain unconstrained.}
\label{fig:freebar_pipeline}
    \label{fig:pipeline} 
\end{figure}

\subsection{CBF-inspired Barrier-Guided Action-Space Projection}

At generation time $t\in[0,1)$, FreeBar first estimates the clean action chunk. 
For flow-based policies, this estimate is given by
\begin{equation}
    \hat{x}_1=x_t+(1-t)v_\theta(x_t,t,c).
\end{equation}
After executing the first $s$ actions from the previous chunk, the residual reference is padded to length $H$, with invalid entries masked out, yielding a fixed-length reference chunk $\bar{A}$. 
Given $\hat{x}_1$ and $\bar{A}$, we define the weighted temporal consistency score
\begin{equation}
    B(\hat{x}_1)
    =
    -\frac{1}{2}
    \sum_{i=0}^{H-1}
    \tilde{w}_i
    \|\hat{x}_{1,i}-\bar{a}_i\|_2^2,
\end{equation}
where $\tilde{w}_i$ jointly encodes reference validity and horizon-wise constraint strength. 
Equivalently, $E_w(\hat{x}_1)=-B(\hat{x}_1)\geq0$ represents the full weighted temporal discrepancy. 
The corresponding action-space guidance direction is
\begin{equation}
    g=\nabla_{\hat{x}_1}B(\hat{x}_1),
    \qquad
    g_i=-\tilde{w}_i(\hat{x}_{1,i}-\bar{a}_i).
\end{equation}
Thus, $g$ directly points the estimated clean action chunk toward the temporally aligned reference. Importantly, it is obtained analytically from action-space differences and requires no backpropagation through the VLA action head.

Inspired by control barrier function (CBF) projection~\cite{ames2019control}, FreeBar minimally corrects the base velocity along $g$:
\begin{equation}
\begin{aligned}
    \tilde{v} &= v_\theta+\lambda g,   
    \text{s.t.}\quad
    g^\top\tilde{v}
    &\geq
    b(t),
    \qquad
    b(t)
    =
    \frac{\alpha E_w(\hat{x}_1)}{\tau(t)},
    \qquad
    \tau(t)=1-t .
\end{aligned}
\end{equation}
The remaining-horizon normalization $\tau(t)$ requires the weighted discrepancy to decay sufficiently rapidly as generation approaches the clean-action endpoint. The resulting one-dimensional QP admits the closed-form solution
\begin{equation}
    \lambda
    =
    \max\left(
    \frac{b(t)-g^\top v_\theta}
         {\|g\|_2^2+\epsilon},
    0
    \right),
    \qquad \epsilon>0.
\end{equation}
For numerical stability under discrete-time inference, we additionally bound the correction magnitude, with an upper bound of $10$ for GR00T, to mitigate finite-step overshoot near the generation endpoint.
We then update the generation state as
\begin{equation}
    x_{t+\Delta t}
    =
    x_t+\Delta t\left(v_\theta+\lambda g\right).
\end{equation}

The projection also admits a Lyapunov-style energy-decay interpretation. Let
$E_w=-B=E_{\mathrm{hard}}+E_{\mathrm{soft}}$ denote the full weighted temporal discrepancy. Under the idealized continuous-time projection, the constraint
\begin{equation}
    g^\top\tilde{v}
    \geq
    \frac{\alpha E_w}{1-t}
\end{equation}
implies
\begin{equation}
    E_w(t)
    \leq
    E_w(0)(1-t)^\alpha.
\end{equation}
Since both components are non-negative and
$E_{\mathrm{hard}}\leq E_w$, decay of the full weighted energy also drives the hard temporal-alignment error toward zero. This should be interpreted as a Lyapunov-style energy-decay property of the idealized continuous-time projection, rather than a formal CBF forward-invariance guarantee. Moreover, for $\alpha>1$, the corresponding barrier target satisfies
$b(t)=O((1-t)^{\alpha-1})$ near the endpoint and therefore remains bounded.

Overall, FreeBar only requires action-space differences, reductions over the action horizon, and a scalar closed-form correction, without backward computation through the VLA model.

\subsection{Progressive Barrier Tightening}

We now analyze how $\tau(t)=1-t$ progressively tightens the constraint. 
Since FreeBar does not differentiate $\hat{x}_1$ through the VLA action head, 
we interpret $g^\top \tilde{v}$ under the same local action-space surrogate used by the projection, 
i.e., as the first-order directional change of $B(\hat{x}_1)$ along the corrected velocity $\tilde{v}$.

For the hard-constrained entries with $\tilde{w}_i=1$, we introduce an auxiliary clean-action trajectory 
$z(t)$ with $\dot{z}(t)=\tilde{v}(t)$, and measure its deviation from the carried-over reference by
\begin{equation}
    E(t)
    =
    -B_{\mathrm{hard}}(z(t))
    =
    \frac{1}{2}
    \sum_{i:\tilde{w}_i=1}
    \|z_i(t)-\bar{a}_i\|_2^2 .
\end{equation}
Under the ideal CBF-style condition 
$g(t)^\top\tilde{v}(t)\geq -\alpha B_{\mathrm{hard}}(z(t))/(1-t)$, 
where $g(t)$ denotes the gradient of $B_{\mathrm{hard}}$ with respect to $z(t)$, 
the deviation energy satisfies
\begin{equation}
    \dot{E}(t)
    \leq
    -\frac{\alpha}{1-t}E(t),
    \qquad
    E(t)
    \leq
    E(0)(1-t)^\alpha .
\end{equation}
Thus, for any $\alpha>0$, $E(t)$ vanishes as $t\to1^{-}$, showing that the hard-constrained portion of the current chunk is progressively tightened toward the carried-over reference under the local surrogate.

\subsection{Horizon-wise Graded Temporal Constraint}

Progressive tightening determines when the constraint becomes strict during the iterative generation; we now specify where it is applied along the prediction horizon. 
Constraining only the immediate overlap avoids boundary discontinuities but may leave later actions inconsistent, whereas enforcing hard consistency over the full horizon can suppress adaptation. 
FreeBar therefore uses a graded horizon mask: proximal actions are hard-aligned, medial actions are softly aligned, and distal actions are left unconstrained for flexibility.

Let $d$ denote the number of control steps elapsed during inference, which directly determines the hard-aligned prefix, and let $h_s$ denote the soft-alignment boundary, with $0\leq d\leq h_s\leq H$. The horizon-wise weight is defined as
\begin{equation}
w_i =
\begin{cases}
1, & 0 \leq i < d, \\
\exp\!\left(-\beta \frac{i-d+1}{h_s-d}\right), & d \leq i < h_s, \\
0, & h_s \leq i < H,
\end{cases}
\end{equation}
where $\beta>0$ controls the decay rate, and the soft region is omitted when $h_s=d$.
The soft boundary $h_s$ controls how far temporal regularization extends beyond the committed prefix and is set to the execution horizon $h$ by default.
To handle invalid aligned references, we use a binary validity indicator $m_i\in\{0,1\}$ and define the effective weight as $\tilde{w}_i=m_iw_i$.
Thus, $\tilde{w}_i$ determines where the constraint is applied along the prediction horizon, while $\tau(t)=1-t$ determines how strongly it is tightened during generation.

The complete FreeBar inference procedure is summarized in Algorithm~\ref{alg:freebar}.

\begin{algorithm}[H]
\caption{FreeBar Inference}
\label{alg:freebar}
\small
\begin{algorithmic}[1]
\REQUIRE VLA action head $v_\theta$, context $c$, reference chunk $\bar{A}$, validity mask $m$, hard length $d$, soft horizon $h_s$, parameters $\alpha,\beta,\epsilon$
\STATE Construct horizon-wise weights $w_i$ and set $\tilde{w}_i \leftarrow m_i w_i$ for $i=0,\dots,H-1$
\STATE Initialize $x_{t_0}\sim\mathcal{N}(0,I)$ with $t_0=0$
\FOR{$k=0,\dots,K-1$}
    \STATE Set generation time $t_k\in[0,1)$ and step size $\Delta t_k=t_{k+1}-t_k$
    \STATE Compute $v_k \leftarrow v_\theta(x_{t_k},t_k,c)$ and $\hat{x}_1 \leftarrow x_{t_k}+(1-t_k)v_k$
    \STATE Compute $B \leftarrow -\frac{1}{2}\sum_{i=0}^{H-1}\tilde{w}_i\|\hat{x}_{1,i}-\bar{a}_i\|_2^2$
    \STATE Compute $g_i \leftarrow -\tilde{w}_i(\hat{x}_{1,i}-\bar{a}_i)$ for $i=0,\dots,H-1$
    \STATE Set $b_k \leftarrow -\alpha B/(1-t_k)$
    \STATE Compute the projection coefficient
    \[
        \lambda
        \leftarrow
        \max\left(
        \frac{
        b_k-g^\top v_k
        }{
        \|g\|_2^2+\epsilon
        },
        0
        \right)
    \]
    \STATE Update $x_{t_{k+1}} \leftarrow x_{t_k}+\Delta t_k(v_k+\lambda g)$
\ENDFOR
\RETURN aligned action chunk $A=x_{t_K}$
\end{algorithmic}
\end{algorithm}

\section{Experiments}
\label{sec:experiments}
To comprehensively evaluate our method, we design our experiments to investigate four key aspects: \textbf{(1)} the improvement of task success rates and execution stability under asynchronous inference in both simulated and real-world environments; \textbf{(2)} the reduction in inference overhead compared to VJP-based baselines like RTC; \textbf{(3)} the necessity of core components, specifically the horizon-wise graded constraints; and \textbf{(4)} the method's robustness across diverse delay conditions, execution horizons, hyperparameters, and base VLA policies.

\begin{wrapfigure}{r}{0.68\textwidth}
    \centering
    \vspace{-2.0cm}
    \includegraphics[width=0.65\textwidth]{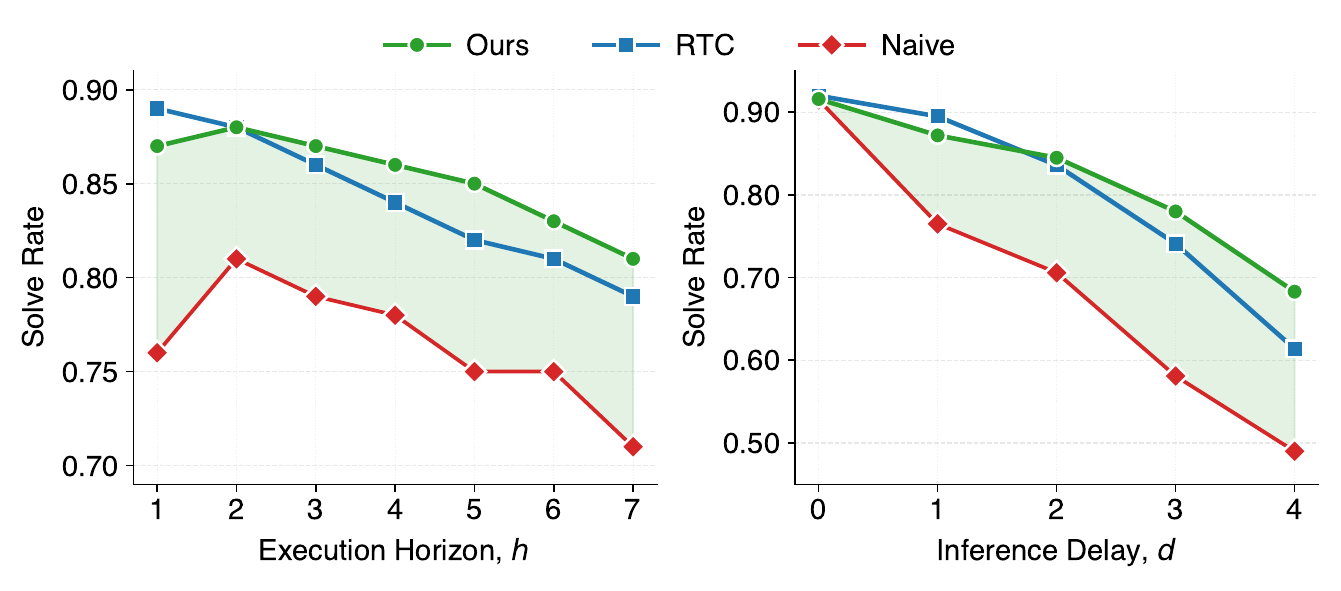}
    \caption{
        \textbf{Left}: Success rates across varying execution horizons, with the inference delay fixed at 1. FreeBar consistently outperforms RTC and yields substantial improvements over the naive baseline. 
        \textbf{Right}: Success rates across varying inference delays, where the execution horizon is set to $h = \max(d, 1)$, with $d$ being the delay. FreeBar maintains superior performance over RTC under high-latency conditions. 
        Each data point is averaged over 100 trials.
        }
    \label{fig:kinetix}
\end{wrapfigure}





\subsection{Comparison Experiments}
\label{sec:compare}
\subsubsection{Evaluation on Kinetix}
\label{sec:kinetix}
\paragraph{Experimental Setup.} We first assess FreeBar using the Kinetix benchmark~\cite{matthews2025kinetix}, following RTC~\cite{black2026real}.
Kinetix provides highly dynamic simulated robotic tasks (e.g., throwing, catching, balancing) that demand real-time reactions and highlight the state-action mismatches caused by the delayed asynchronous execution.
We train action-chunking flow policies with a prediction horizon of $H = 8$ and a 4-layer MLP-Mixer architecture for 32 epochs.
We measure average success rates across the same 12 tasks, testing simulated delays $d \in [0, 4]$ and varying the execution horizon $h$ to analyze the effects of temporal chunk alignment.

\paragraph{Results.}
Figure~\ref{fig:kinetix} summarizes the Kinetix results under varied execution horizons and inference delays.
In the horizon sweep, RTC and FreeBar both consistently outperform naive asynchronous execution.
While RTC is slightly stronger at the shortest horizon, FreeBar remains more stable as the execution horizon increases (e.g., at $h=7$, FreeBar achieves $81.0\%$ vs. RTC's $79.0\%$ and naive's $71.0\%$).
In the delay sweep, the performance of naive execution deteriorates rapidly as latency increases, with its success rate dropping from $91.5\%$ at $d=0$ to $49.0\%$ at $d=4$.
RTC improves robustness under small delays, but FreeBar performs best under larger delays, reaching a success rate of $68.3\%$ at $d=4$ compared to $61.4\%$ for RTC.
These results indicate that FreeBar effectively mitigates prediction-execution misalignment under severe inference latency.

\subsubsection{Evaluation on RoboCasa}
\label{sec:robocasa}
\paragraph{Experimental Setup.} 
We evaluate FreeBar on RoboCasa~\cite{robocasa365} across 2 VLA backbones including GR00T~\cite{gr00tn1_2025} and \pio ~\cite{black2026pi0visionlanguageactionflowmodel} on 11 long-horizon manipulation tasks.
For GR00T, we use a checkpoint that was pretrained for 80k steps, followed by target-task fine-tuning for 60k steps.
For \pio, we train the model for 75k steps with a batch size of 64 and evaluate it in pretraining scenes.
We compare FreeBar against naive asynchronous execution, RTC, and a non-horizon-wise CBF ablation that removes horizon-wise graded temporal constraints.
Additionally, we conduct timing experiments on a single RTX A100 GPU using GR00T as the base policy, with the prediction horizon at $H = 16$, the inference delay at $d=3$, the execution horizon at $h=6$, and the denoising steps set to 4.


\begin{table}[htbp]
    \centering
    \caption{\textbf{Left:} RoboCasa results (\%) with GR00T and \pio{} as base policies.
    \textbf{Right:} Hyperparameter robustness across tested configurations, where higher mean performance and lower standard deviation indicate stronger and more stable robustness.}
    \vspace{-0cm}
    \label{tab:combined_tables}
    \small
    \setlength{\tabcolsep}{3pt}
    \begin{minipage}[t]{0.5\textwidth}
        \centering
        \begin{tabular}[t]{@{}lcccccc@{}}
        \toprule
        & \multicolumn{3}{c}{\textbf{Base Policy: GR00T}} & \multicolumn{3}{c}{\textbf{Base Policy: \pio}} \\
        \cmidrule(lr){2-4} \cmidrule(lr){5-7}
        Method & $d=1$ & $d=2$ & $d=3$ & $d=1$ & $d=2$ & $d=3$ \\
        \midrule
        Naive   & 68.0 & 69.6 & 68.6 & 33.8 & 32.7 & 31.2 \\
        RTC     & 69.6 & 70.5 & 70.0 & 31.3 & 32.7 & 31.7 \\
        CBF     & 68.5 & 69.8 & 68.0 & 31.3 & 33.3 & 34.0 \\
        FreeBar & \textbf{72.8} & \textbf{72.3} & \textbf{71.4} & \textbf{36.0} & \textbf{36.5} & \textbf{36.3} \\
        \bottomrule
        \end{tabular}
    \end{minipage}\hfill
    \begin{minipage}[t]{0.48\textwidth}
        \centering
        \begin{tabular}[t]{@{}lcc@{}}
        \toprule
        Task & RTC & FreeBar \\
        \midrule
        RoboCasa (sim., \%) $^\ast$ & 65.93 $\pm$ 6.55 & \textbf{69.63 $\pm$ 1.77} \\
        \midrule
        Conv.\ Belt (\%) $^\dagger$ & 55.53 $\pm$ 7.70 & \textbf{74.67 $\pm$ 6.04} \\
        Conv.\ Belt (\%) $^\ddagger$ & 59.33 $\pm$ 9.25 & \textbf{75.99 $\pm$ 4.35} \\
        Whack-a-Mole (hits) $^\dagger$ & 7.80 $\pm$ 1.92 & \textbf{10.40 $\pm$ 1.14} \\
        \bottomrule
        \end{tabular}
        \vspace{0.15cm} 
        \raggedright
        \scriptsize
        \textit{Note:} Base policies used are GR00T ($^\ast$), \pio~ ($^\dagger$), and \pihalf~ ($^\ddagger$).
    \end{minipage}
    \vspace{-0.7cm}
\end{table}

\begin{wrapfigure}{R}{0.67\textwidth}
    \centering
    \begin{minipage}[t]{0.52\linewidth}
        \centering
        \includegraphics[width=\linewidth]{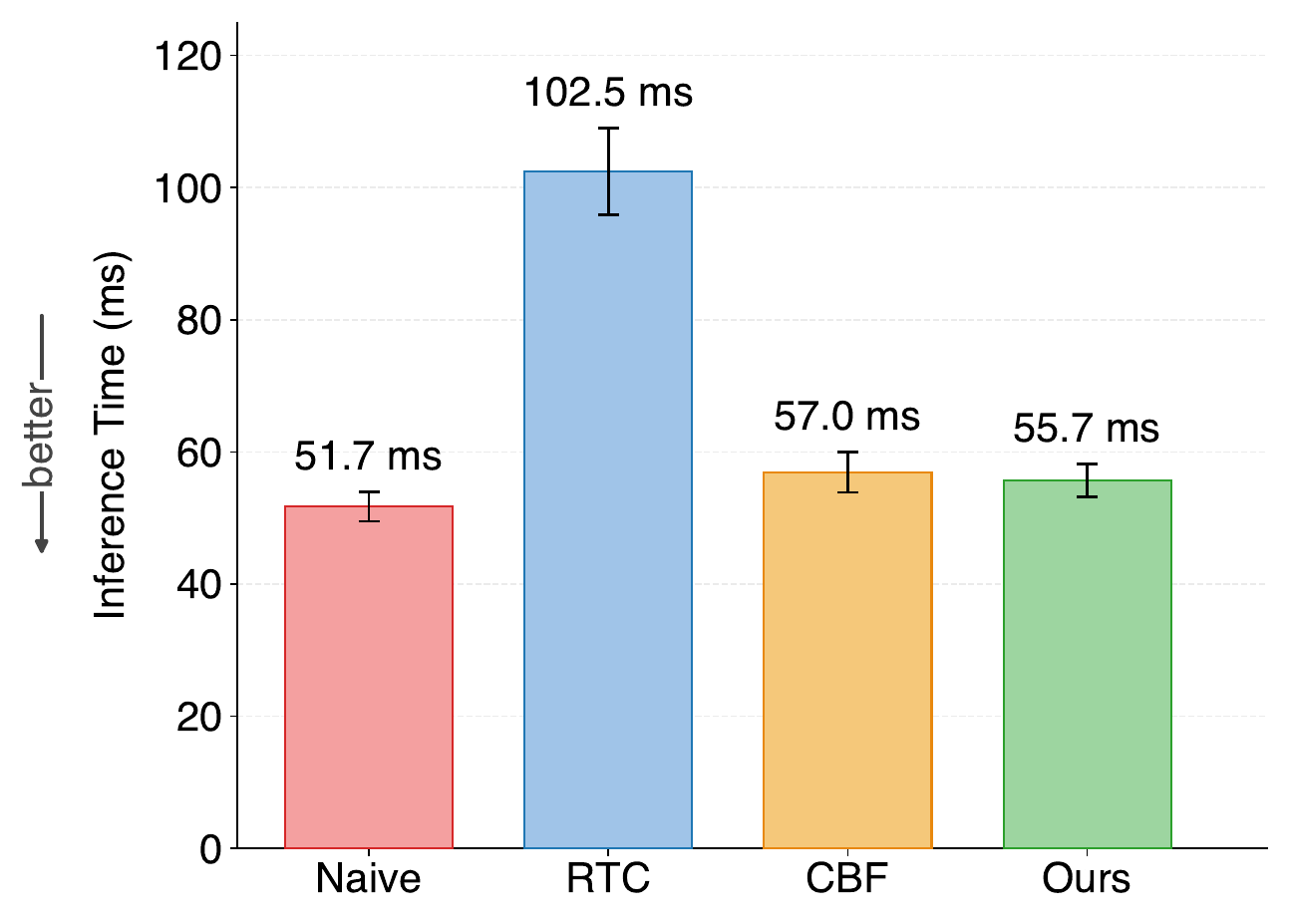}
    \end{minipage}%
    \hfill
    \begin{minipage}[t]{0.48\linewidth}
        \centering
        \includegraphics[width=\linewidth]{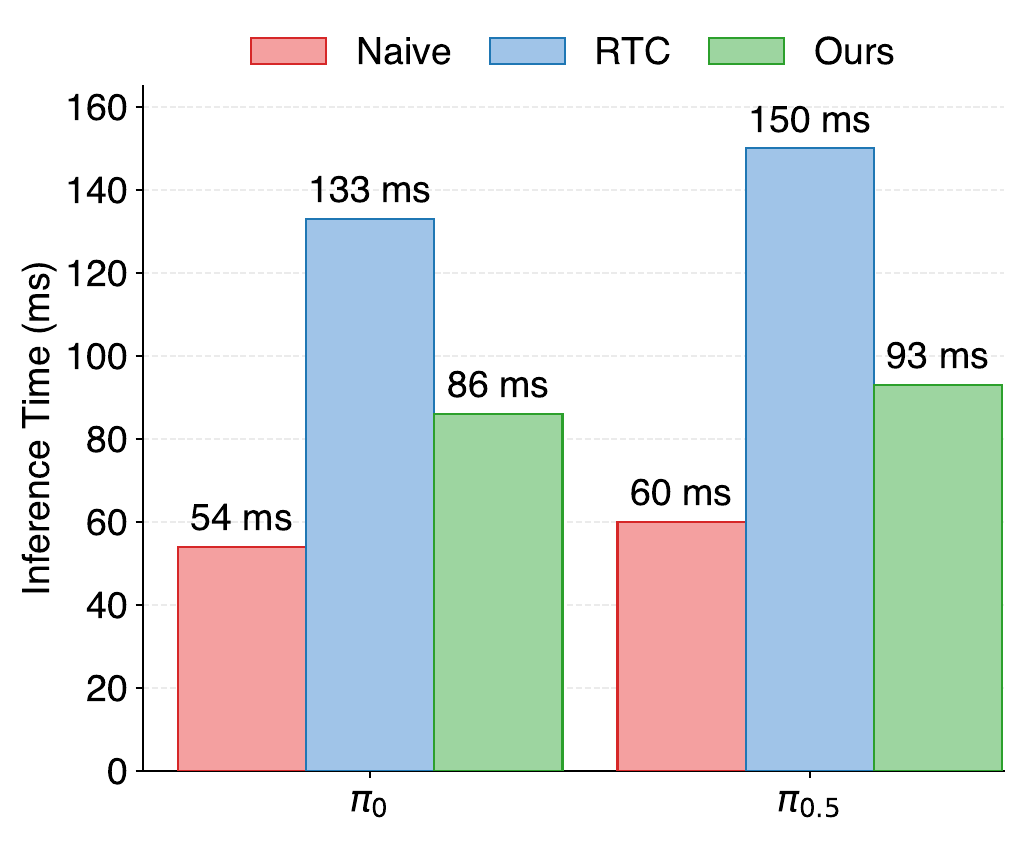}
    \end{minipage}
    \caption{
        \textbf{Left}: Simulation inference time on RoboCasa with GR00T. FreeBar incurs minimal overhead, comparable to non-horizon-wise CBF, while significantly outperforming RTC.
        \textbf{Right}: Real-world inference time on the conveyor-belt task with \pio~ and \pihalf~. FreeBar maintains substantially lower latency than RTC.
    }
    \label{fig:efficiency}
\end{wrapfigure}

\paragraph{Results.} Table~\ref{tab:combined_tables} (left) reports RoboCasa results with GR00T and \pio~ as base policies. Across both backbones and all delay settings, FreeBar consistently achieves the highest success rate.
For GR00T, FreeBar outperforms all baselines (naive, RTC, CBF) across all delay levels (e.g., $71.4\%$ at $d=3$).
For \pio, FreeBar still improves performance even when other baselines fail.

Regarding the computational efficiency across methods (the left panel in Figure~\ref{fig:efficiency}), RTC suffers from heavy inference latency due to repeated VJP-based guidance.
Conversely, FreeBar relies solely on action-space discrepancy computation and light-weight closed-form projection, which incur minimal overhead ($+7.7\%$), comparable to CBF ($+10.1\%$) but significantly lower than RTC ($+98.0\%$).

\subsection{Ablation Experiments \& Hyperparameter Robustness} 
\label{sec:ablation_hyperparamter}
\paragraph{Ablation on horizon-wise graded constraints.}
Table~\ref{tab:combined_tables} (left) ablates the horizon-wise graded temporal constraint.
The non-graded variant, denoted as \textsc{CBF}, applies uniform barrier constraint over the alignment horizon using the same closed-form correction, which yields marginal gains and can underperforms naive asynchronous execution, such as GR00T at $d=3$ and \pio at $d=1$.
This suggests fixed consistency constraints may over-restrict uncommitted actions and introduce abrupt boundary transitions, causing jerky execution.
Conversely, FreeBar’s horizon-wise grading balances near-term alignment with long-horizon adaptability, achieving consistently stronger performance.

\paragraph{Comparison with lightweight temporal baselines.}
We further compare FreeBar with lightweight temporal-consistency baselines,
including Temporal Ensembling, Action-Space Smoothing, and
Chunk-Boundary Interpolation, to examine whether its gains simply arise
from generic smoothing. As shown in Table~\ref{tab:lightweight_baselines},
FreeBar consistently outperforms these alternatives on both Kinetix and
RoboCasa, with particularly stronger robustness under increased inference
delay on Kinetix. These results suggest that generic averaging or
interpolation is insufficient for asynchronous chunk alignment.
Instead, FreeBar performs violation-triggered minimum-change correction
during generation while preserving distal-action flexibility through
horizon-wise graded constraints.

\begin{table}[t]
\centering
\caption{
Comparison with lightweight temporal-consistency baselines.
FreeBar consistently outperforms generic smoothing and alignment
strategies under varying inference delays $d$.
}
\label{tab:lightweight_baselines}

\small
\setlength{\tabcolsep}{6pt}
\renewcommand{\arraystretch}{1.08}

\begin{tabular}{@{}llccc@{}}
\toprule
\textbf{Benchmark} & \textbf{Method}
& $\mathbf{d=1}$ & $\mathbf{d=2}$ & $\mathbf{d=3}$ \\
\midrule

\multirow{2}{*}{Kinetix}
& Temporal Ensembling
& 41.8 & 17.2 & 3.3 \\
& FreeBar
& \textbf{87.2} & \textbf{84.5} & \textbf{78.0} \\

\midrule

\multirow{4}{*}{RoboCasa}
& Action-Space Smoothing
& 70.7 & 68.9 & 69.7 \\
& Chunk-Boundary Interpolation
& 65.5 & 64.2 & 69.7 \\
& Uniform Hard Alignment
& 69.3 & 68.9 & 67.9 \\
& FreeBar
& \textbf{72.8} & \textbf{72.3} & \textbf{71.4} \\

\bottomrule
\end{tabular}
\end{table}

\paragraph{Hyperparameter robustness.}

Table~\ref{tab:combined_tables} (right) reports the mean and standard deviation of task performance when sweeping correction strengths (RTC's $\beta$ and FreeBar's $\alpha$) across $[0.1, 50]$ (see full discrete search space in Appendix).
Since $\alpha$ and $\beta$ operate on different scales, we assess robustness at the method level.
Across all settings, FreeBar achieves higher mean performance and lower standard deviation than RTC.
In simulation, FreeBar's standard deviation is significantly lower, indicating closed-form projection is less sensitive to the correction strength than VJP-based guidance.
This advantage extends to real-world deployments.
On the conveyor-belt task, FreeBar's mean success rate exceeds RTC's by $17$--$19$ percentage points.
Similarly, on the whack-a-mole task, its hit rate increases by 33\%, while its standard deviation decreases by 41\%.
These results confirm FreeBar's barrier-constrained projection is inherently robust, reducing the need for task-specific tuning.

\subsection{Real-World Experiments}
\label{sec:realworld}

\begin{figure}[h] 
    \centering
    \includegraphics[width=1\textwidth]{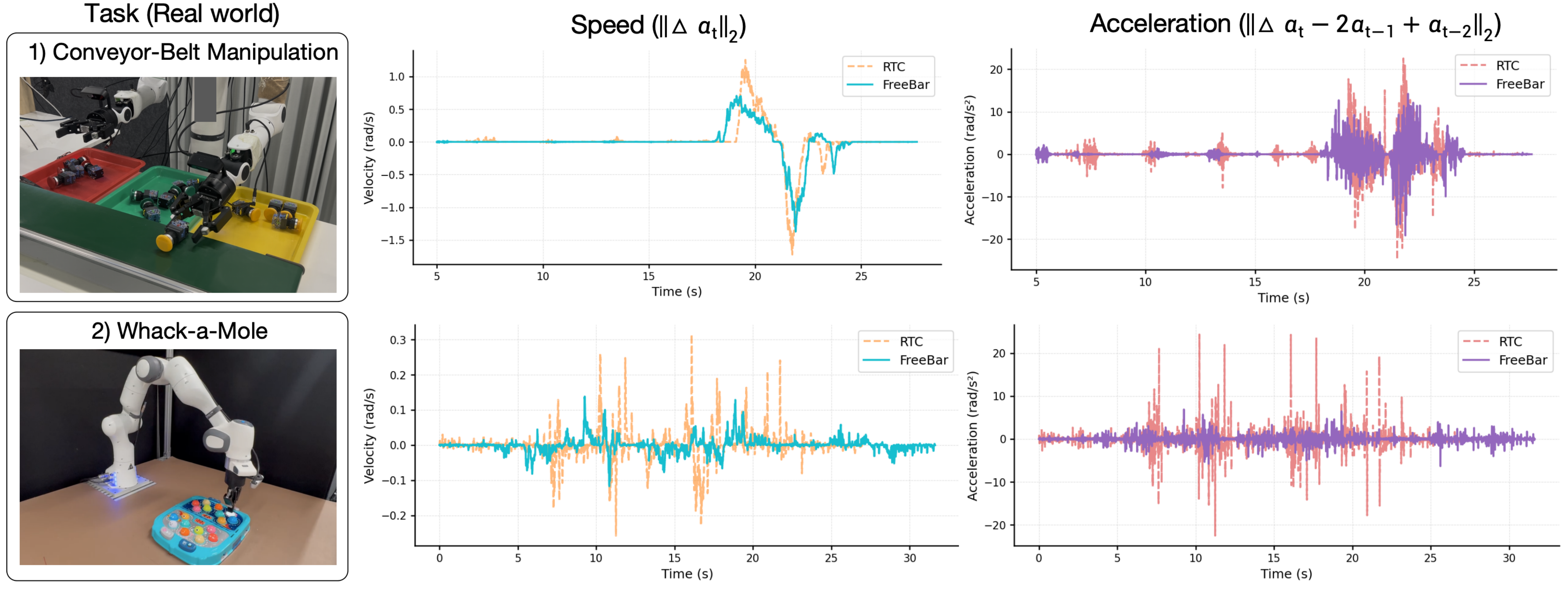} 
    \captionsetup{skip=1pt}
    \caption{\textbf{Real-world kinematic smoothness comparison between RTC and FreeBar.}
    \textbf{Top row}: Conveyor-belt manipulation (TianGong).
    \textbf{Bottom row}: Whack-a-mole (Franka Emika robot).
    \textbf{Middle \& Right}: Wrist joint velocity and acceleration.
    During active manipulation, RTC exhibits remarkable oscillations in both metrics, whereas FreeBar yields substantially smoother profiles.}
    \label{fig:real} 
\end{figure}

\paragraph{Experimental Setup.} 
We deploy FreeBar with \pio{} and \pihalf{} on a humanoid robot equipped with dual 7-DOF arms (TianGong) and a Franka Emika robot (7-DOF arm).
We conduct all experiments at a control frequency of 30 Hz using a single NVIDIA RTX 4090 GPU.
Figure~\ref{fig:real} (left) shows two dynamic tasks: \textbf{Conveyor-Belt Manipulation (TianGong)}, which involves grasping and sorting moving buttons to test temporal alignment between stale observations and delayed actions; and \textbf{Whack-a-Mole (Franka Emika robot)}, which requires rapid responses to abruptly changing targets to evaluate real-time reactive control.
Across 30 rollouts per datapoint, we compare FreeBar against naive asynchronous execution and RTC, reporting the success rate of the conveyor-belt task as well as the total number of successful hits within 30\,s for whack-a-mole task.

\paragraph{Task Performance.}
Table~\ref{tab:main_results_vertical} shows that FreeBar outperforms naive asynchronous execution and RTC across most settings and backbones on both tasks.
The gain over the naive baseline confirms executing the latest available chunk fails in continuously evolving environments.
Furthermore, FreeBar's superiority over RTC suggests that closed-form barrier projection is more stable for physical deployment, where latency and actuation noise amplify the instability of VJP-based guidance.

\begin{table}[h] 
\centering
\caption{Real-world manipulation results at a fixed inference delay ($d=2$) across varied execution horizons ($h$). We report Conveyor-Belt Manipulation success rates (\%) and total Whack-a-Mole hits within 30\,s. Our method achieves the highest performance across almost all tested settings.}
\label{tab:main_results_vertical}
\vspace{0.1cm}
\resizebox{\textwidth}{!}{
\begin{tabular}{l | cccc | cccc | cccc}
\toprule
\multirow{2}{*}{Method} & \multicolumn{4}{c|}{Conveyor Belt (\pio)} & \multicolumn{4}{c|}{Conveyor Belt (\pihalf)} & \multicolumn{4}{c}{Whack-a-Mole (\pio)} \\
& $h=6$ & $h=8$ & $h=10$ & $h=12$ & $h=6$ & $h=8$ & $h=10$ & $h=12$ & $h=6$ & $h=8$ & $h=10$ & $h=12$ \\
\midrule
Naive & 56.67 & 70.00 & 66.67 & 63.33 & 63.33 & 66.67 & 76.67 & 63.33 & 2.03 & 5.02 & 7.05 & 5.98 \\
RTC   & 73.33 & 83.33 & 80.00 & \textbf{76.67} & 76.67 & 80.00 & 83.33 & 80.00 & \textbf{10.05} & 9.98 & 10.02 & 8.01 \\
\midrule 
Ours  & \textbf{90.00} & \textbf{86.67} & \textbf{86.67} & \textbf{76.67} & \textbf{93.33} & \textbf{90.00} & \textbf{90.00} & \textbf{83.33} & 9.61 & \textbf{13.04} & \textbf{12.58} & \textbf{10.62} \\
\bottomrule
\end{tabular}
}
\end{table}

\paragraph{Inference Efficiency.}
Figure~\ref{fig:efficiency} (right) compares real-world computational overhead on the conveyor-belt task for \pio~and~\pihalf~(RTX 4090, $H=50$, $d=3$, $h=5$, 10 denoising steps).
FreeBar requires only $86$ ms for \pio~and $93$ ms for \pihalf~(vs.\ $133$ and $150$ ms for RTC).
This minimal overhead preserves latency budgets, enabling asynchronous closed-loop control at 30~Hz.

\paragraph{Kinematic Smoothness.}
The middle and right panels of Figure~\ref{fig:real} compare wrist joint velocity and acceleration profiles of FreeBar and RTC, omitting the severely misaligned naive baseline. 
As shown, RTC exhibits high-frequency oscillations in both metrics during active manipulation, indicating VJP-based guidance introduces unstable corrections at chunk boundaries.
Conversely, FreeBar yields smoother kinematic profiles, since its closed-form barrier projection enforces temporal continuity at chunk transitions, suppressing disruptive artifacts and improving physical control stability.

\section{Conclusion and Limitation}
\label{sec:conclusion}
We presented \textit{FreeBar}, a training-free real-time inference plugin that mitigates prediction-execution misalignment in asynchronous VLA execution via barrier-constrained closed-form action-space projection, progressively aligning new chunks with the ongoing execution stream. Extensive evaluations demonstrate that FreeBar robustly enhances task success rates with negligible inference overhead compared to existing baselines. These findings prove that lightweight, barrier-guided alignment is highly practical for latency-robust VLA deployment.

\textbf{Limitations}. FreeBar's performance depends on the reliability of the carry-over reference. Additionally, forced alignment on proximal part of action chunk reduces reactivity in highly dynamic scenarios. Future work may explore confidence-aware reference selection and adaptive barrier scheduling for stronger training-free alignment.
\acknowledgments{The authors would like to thank the team at the Beijing Innovation Center of Humanoid Robotics for their valuable contributions and support.}
\bibliography{main}  
\clearpage
\appendix 
\setcounter{table}{0}
\renewcommand{\thetable}{A\arabic{table}} 
\setcounter{figure}{0}
\renewcommand{\thefigure}{A\arabic{figure}} 
\setcounter{equation}{0}
\renewcommand{\theequation}{A\arabic{equation}} 

\section{Implementation and Experimental Analysis}
\label{sec:app_exp_details}
Our code is available at: \url{https://github.com/NamelessCrew/FreeBar}\\
This section provides additional details for the experimental tasks and evaluation protocols used in the main paper. We include visual illustrations and task descriptions for the Kinetix and RoboCasa benchmarks, describe the two real-world dynamic manipulation tasks, and report the complete hyperparameter-robustness results. Overall, our evaluation covers a broad range of task structures, including physics-based dynamic control and household manipulation, single-arm reactive manipulation, and dual-arm dynamic sorting. These settings are intentionally selected to stress-test real-time asynchronous execution under stale observations, delayed action generation, moving targets, and chunk-boundary misalignment.

\subsection{Task Suite Overview}
\label{sec:app_task_suite_overview}

Table~\ref{tab:app_task_suite_summary} summarizes the task suites used in our experiments. Kinetix evaluates real-time physical interaction in highly dynamic 2D environments. RoboCasa evaluates complex manipulation in visually rich household scenes. The real-world experiments further test physical deployment with moving objects and abrupt target changes. Together, these benchmarks cover both simulated and real robotic settings, single-arm and dual-arm embodiments, static and dynamic scenes, and both short-horizon reactive control and long-horizon manipulation.

\begin{table}[h!]
\centering
\caption{
Summary of task suites used in our evaluation. The selected tasks cover diverse embodiments, visual domains, control frequencies, and dynamic interaction patterns.
}
\label{tab:app_task_suite_summary}
\small 
\setlength{\tabcolsep}{4pt}
\renewcommand{\arraystretch}{1.2} 
\begin{tabularx}{\linewidth}{
    >{\hsize=0.8\hsize\raggedright\arraybackslash}X  
    c                                                
    >{\hsize=0.9\hsize\raggedright\arraybackslash}X  
    >{\hsize=0.8\hsize\raggedright\arraybackslash}X  
    >{\hsize=1.5\hsize\raggedright\arraybackslash}X  
}
\toprule
\textbf{Suite} & \textbf{\# Tasks} & \textbf{Embodiment / Policy} & \textbf{Domain} & \textbf{Main Challenge} \\
\midrule
Kinetix & 12 & Flow policy with action chunking & Physics-based simulation & Fast reaction, contact-rich dynamics, delayed execution \\
\midrule 
RoboCasa & 11 & GR00T, $\pi_0$ & Household manipulation simulation & Complex manipulation, visual variation, chunk alignment \\
\midrule 
Conveyor-Belt Manipulation & 1 & Dual-arm humanoid, $\pi_0$ / $\pi_{0.5}$ & Real world & Sorting moving objects under stale observations \\
\midrule 
Whack-a-Mole & 1 & Franka Emika, $\pi_0$ & Real world & Abrupt target changes and reactive striking \\
\bottomrule
\end{tabularx}
\end{table}

\subsection{Kinetix Tasks}
\label{sec:app_kinetix_tasks}

Kinetix provides highly dynamic physics-based control tasks that require real-time interaction with moving objects. All tasks use force-based control in 2D rigid-body physics, where the goal is to guide a green object to touch a blue target without contacting red obstacles. These tasks are particularly suitable for evaluating asynchronous VLA execution because delayed inference directly causes state-action mismatch: the object configuration observed by the policy may have already changed by the time the generated action chunk is executed. We use the same 12 Kinetix tasks as in the main evaluation and report average solve rates across 100 trials per configuration.

\begin{figure}[b]
\centering
\begin{subfigure}{0.30\linewidth}
    \centering
    \includegraphics[width=\linewidth]{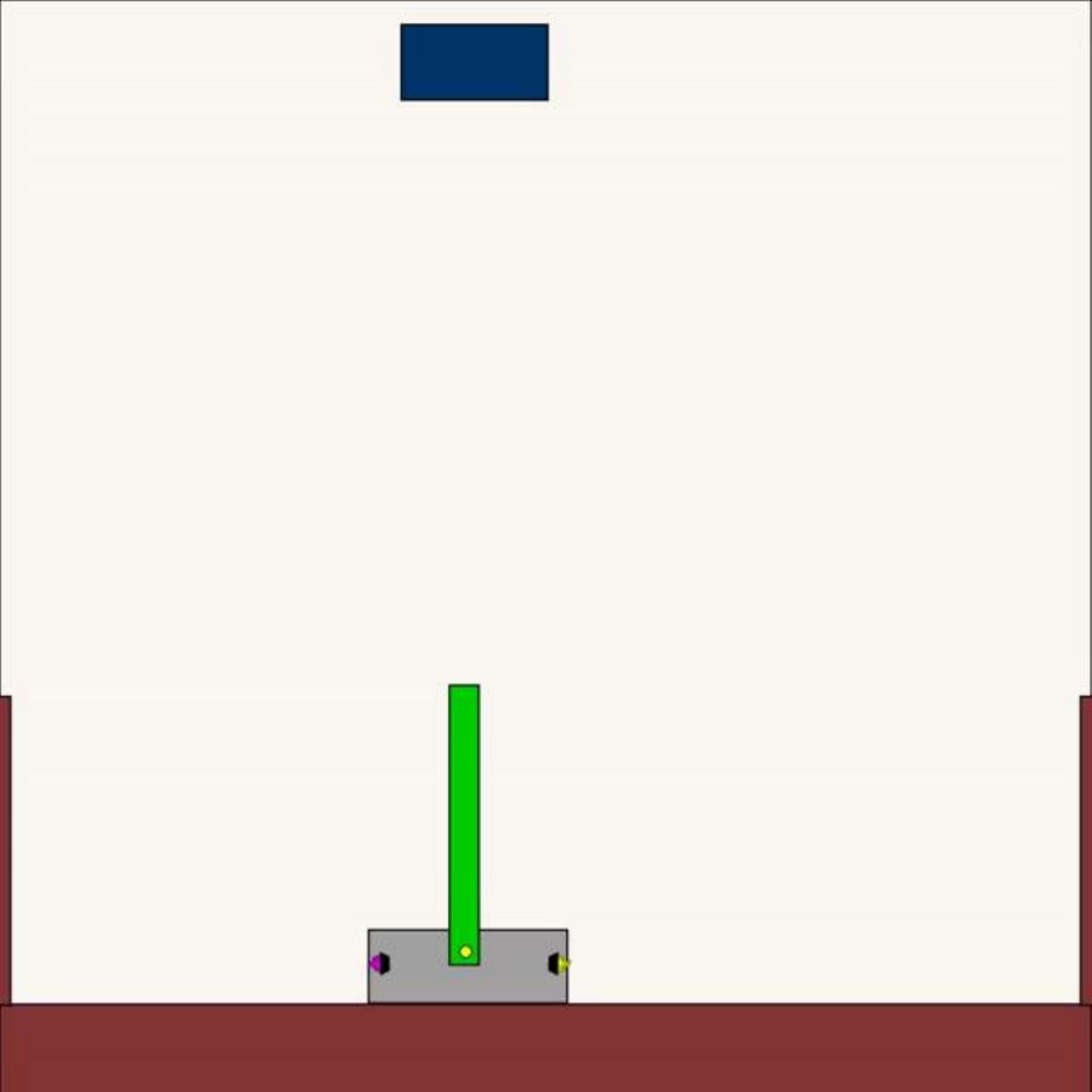}
    \caption{\texttt{cartpole\_thrust}}
\end{subfigure}
\hfill
\begin{subfigure}{0.30\linewidth}
    \centering
    \includegraphics[width=\linewidth]{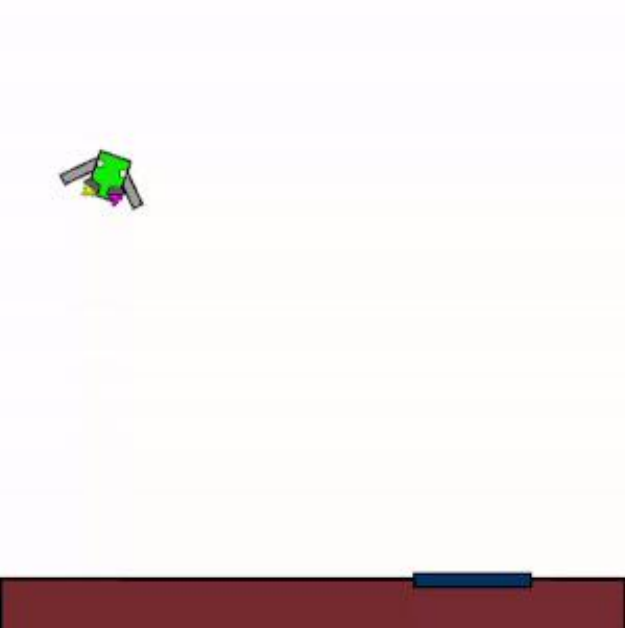}
    \caption{\texttt{hard\_lunar\_lander}}
\end{subfigure}
\hfill
\begin{subfigure}{0.30\linewidth}
    \centering
    \includegraphics[width=\linewidth]{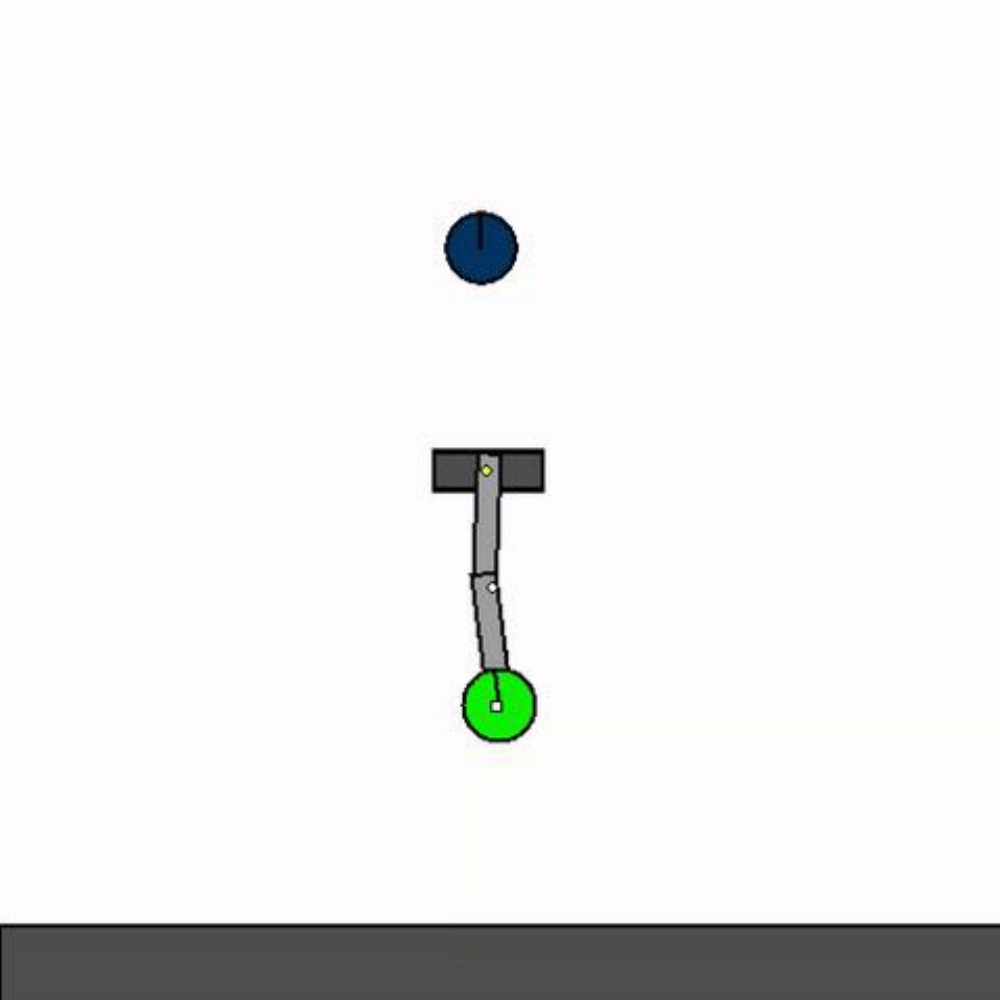}
    \caption{\texttt{catcher\_v3}}
\end{subfigure}

\vspace{0.35em}

\begin{subfigure}{0.30\linewidth}
    \centering
    \includegraphics[width=\linewidth]{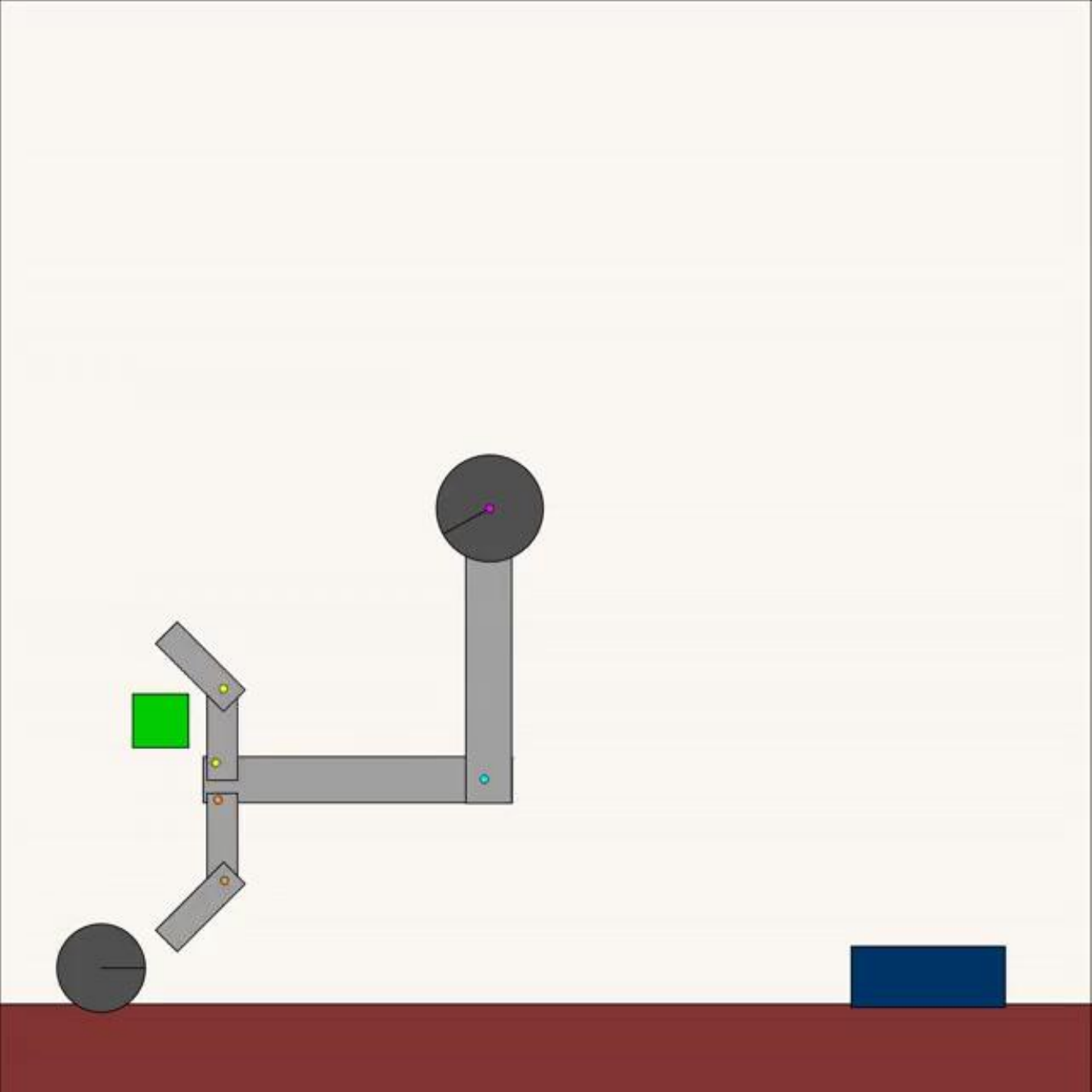}
    \caption{\texttt{grasp\_easy}}
\end{subfigure}
\hfill
\begin{subfigure}{0.30\linewidth}
    \centering
    \includegraphics[width=\linewidth]{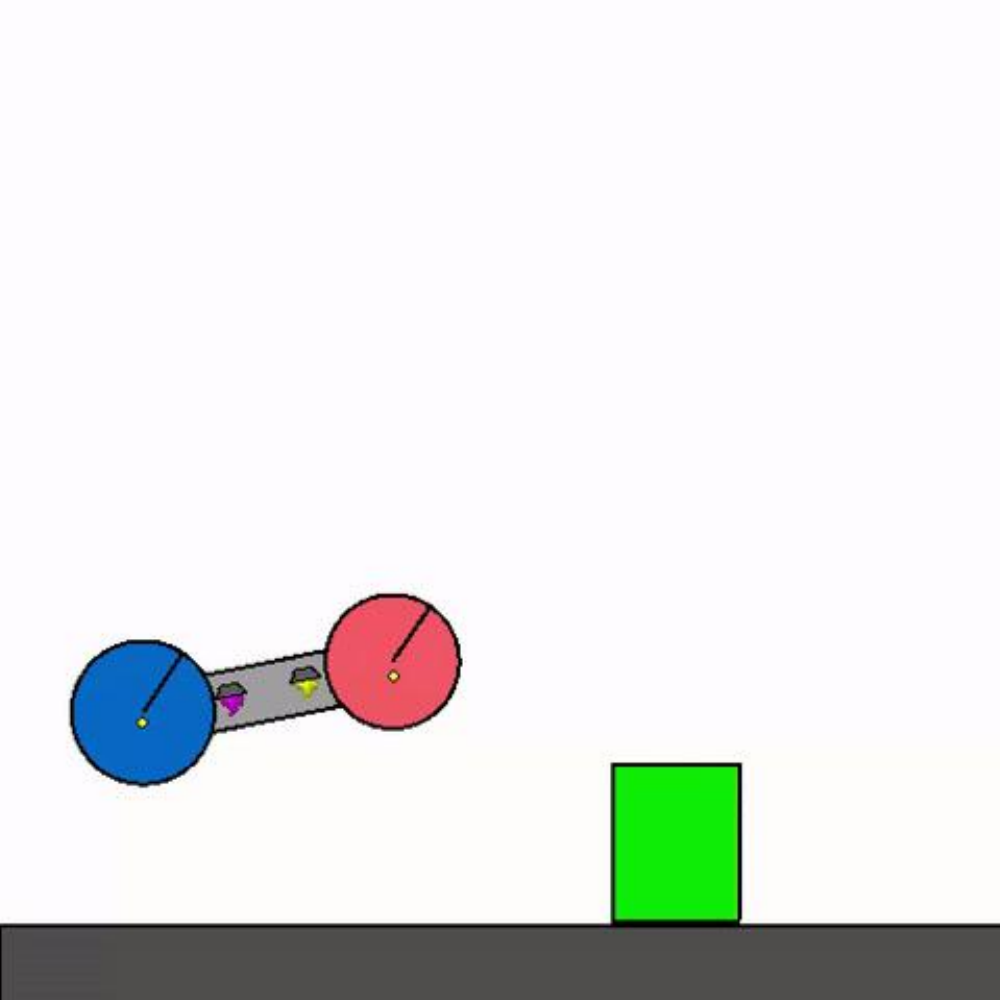}
    \caption{\texttt{catapult}}
\end{subfigure}
\hfill
\begin{subfigure}{0.30\linewidth}
    \centering
    \includegraphics[width=\linewidth]{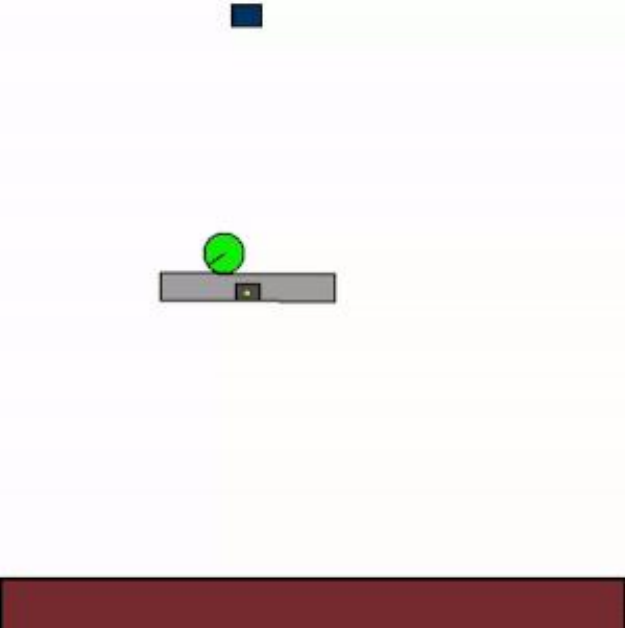}
    \caption{\texttt{trampoline}}
\end{subfigure}
\caption{
Representative Kinetix tasks from our evaluation suite. The six environments shown cover classic balance control (a, b), reactive manipulation (c, d), and dynamic projectile/locomotion tasks (e, f). In each task the agent must guide a green object to touch a blue target without contacting red obstacles using force-based control. See Table~\ref{tab:app_kinetix_task_descriptions} for the full list of all 12 tasks.
}
\label{fig:app_kinetix_tasks}
\end{figure}

Figure~\ref{fig:app_kinetix_tasks} visualizes the 12 Kinetix tasks. Each task requires the policy to generate temporally consistent action chunks while reacting to changing object states. The task set includes diverse dynamic patterns such as reaching, pushing, throwing, catching, balancing, and sequential object interaction. This diversity allows us to evaluate whether FreeBar improves chunk-boundary consistency without over-constraining future actions.

\clearpage

\begin{table}[t]
\centering
\caption{
Description of the 12 Kinetix tasks used in our evaluation.
}
\label{tab:app_kinetix_task_descriptions}
\setlength{\tabcolsep}{4pt}
\renewcommand{\arraystretch}{1.05}
\resizebox{\linewidth}{!}{
\begin{tabular}{clp{0.62\linewidth}}
\toprule
\textbf{ID} & \textbf{Task} & \textbf{Dynamic property evaluated} \\
\midrule
1  & \texttt{car\_launch}        & Launching a vehicle toward a target; requires precise force magnitude and timing under delayed execution. \\
2  & \texttt{cartpole\_thrust}   & Balancing an inverted pendulum with active thrust; demands continuous real-time correction to maintain stability. \\
3  & \texttt{catapult}           & Catapulting an object to a target; actions have irreversible delayed consequences once the projectile is released. \\
4  & \texttt{catcher\_v3}        & Catching a moving object mid-flight; requires rapid reaction to a moving target whose position changes during inference. \\
5  & \texttt{chain\_lander}      & Landing a chain-linked articulated body; coordinated joint control under kinematic constraints and gravity. \\
6  & \texttt{grasp\_easy}        & Grasping an object in a dynamic scene; requires stable interaction as the object state evolves during inference. \\
7  & \texttt{h17\_unicycle}      & Balancing a unicycle; tests the trade-off between proximal chunk continuity and distal action flexibility. \\
8  & \texttt{hard\_lunar\_lander} & A challenging lunar lander variant; fast-changing contact dynamics where over-constraining future actions degrades performance. \\
9  & \texttt{mjc\_half\_cheetah}  & High-speed cheetah locomotion; evaluates robustness to accumulated state-action mismatch across multiple chunks. \\
10 & \texttt{mjc\_swimmer}       & Multi-joint swimmer locomotion; requires real-time correction under continuous body-fluid interaction. \\
11 & \texttt{mjc\_walker}        & Bipedal walking; tests dynamic task completion when action chunks are generated with non-negligible delay. \\
12 & \texttt{trampoline}         & Trampoline jumping and balancing; demands temporally consistent responses to rapid state transitions. \\
\bottomrule
\end{tabular}
}
\end{table}


\subsection{RoboCasa Tasks}
\label{sec:app_robocasa_tasks}

We further evaluate FreeBar on 11 RoboCasa tasks. Compared to Kinetix, the RoboCasa benchmarks feature greater visual complexity and align more closely with practical real-world applications. They require the policy to interpret rich scene observations, follow language instructions, and execute temporally extended manipulation behaviors. We use two VLA backbones, GR00T and $\pi_0$, to verify that FreeBar is not tied to a specific model architecture. Notably, GR00T exhibits significantly higher base success rates than $\pi_0$ in our evaluation. This performance gap is expected given their distinct training regimens: our GR00T checkpoint was pretrained on over 2,000 hours of data across 300 tasks (comprising 482 hours of human data and 1,615 hours of synthetic MimicGen data) and explicitly fine-tuned on an "Atomic-Seen" split of 18 tasks, which fully encompasses our 11 evaluation tasks. In contrast, the $\pi_0$ checkpoint relies solely on broad pretraining without task-specific fine-tuning. By successfully enhancing both a highly domain-optimized policy (GR00T) and a zero-shot generalized policy ($\pi_0$), we demonstrate that FreeBar's effectiveness is robust and broadly applicable, regardless of the underlying policy's initial proficiency.

\begin{figure}[h!]
\centering
\begin{subfigure}{0.325\linewidth}
    \centering
    \includegraphics[width=0.7\linewidth]{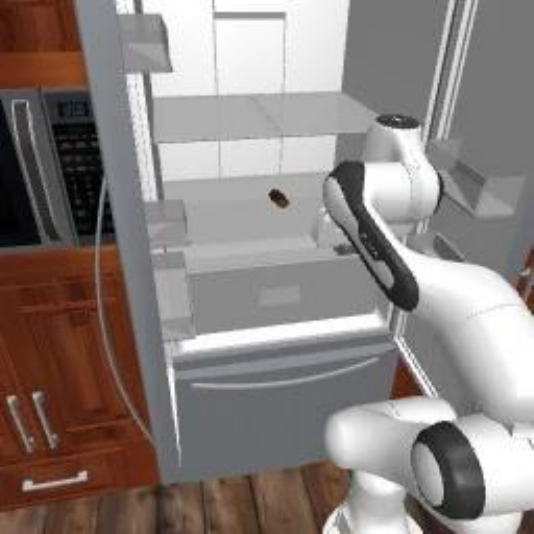}
    \caption{CloseFridge}
\end{subfigure}
\begin{subfigure}{0.325\linewidth}
    \centering
    \includegraphics[width=0.7\linewidth]{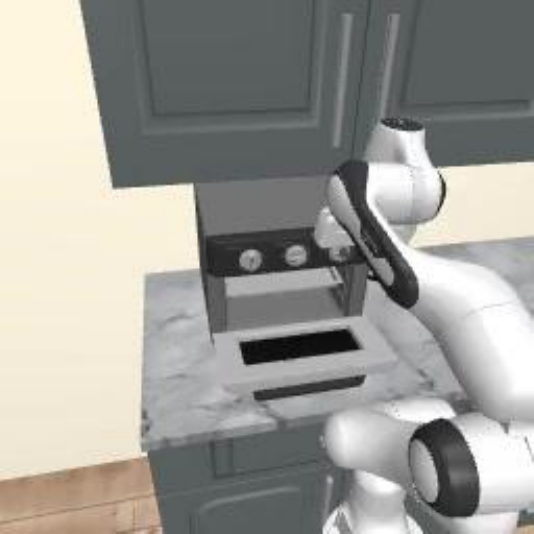}
    \caption{CloseToasterOvenDoor}
\end{subfigure}
\begin{subfigure}{0.325\linewidth}
    \centering
    \includegraphics[width=0.7\linewidth]{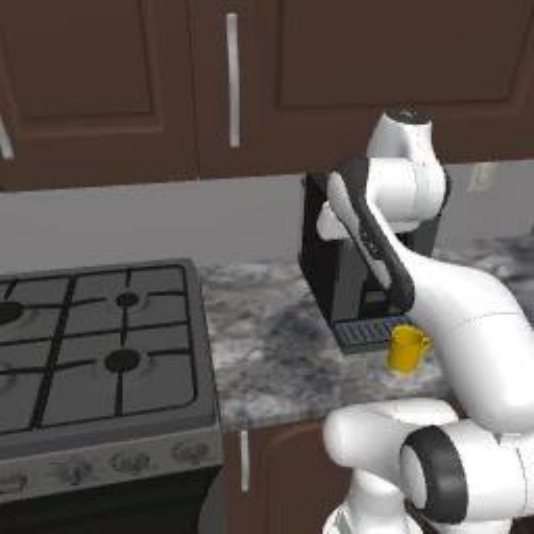}
    \caption{CoffeeSetupMug}
\end{subfigure}

\vspace{0.35em}

\begin{subfigure}{0.325\linewidth}
    \centering
    \includegraphics[width=0.7\linewidth]{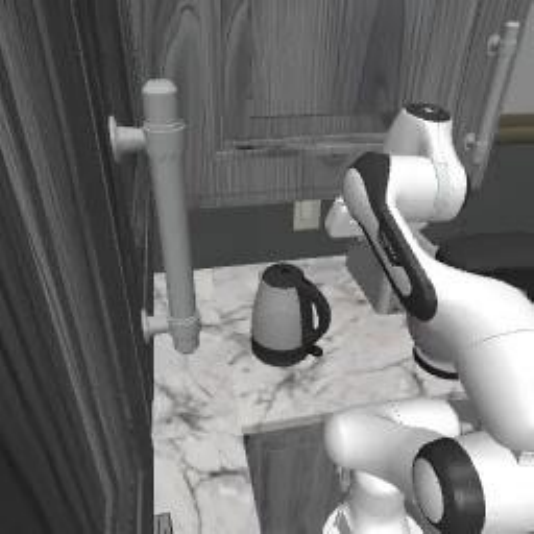}
    \caption{NavigateKitchen}
\end{subfigure}
\begin{subfigure}{0.325\linewidth}
    \centering
    \includegraphics[width=0.7\linewidth]{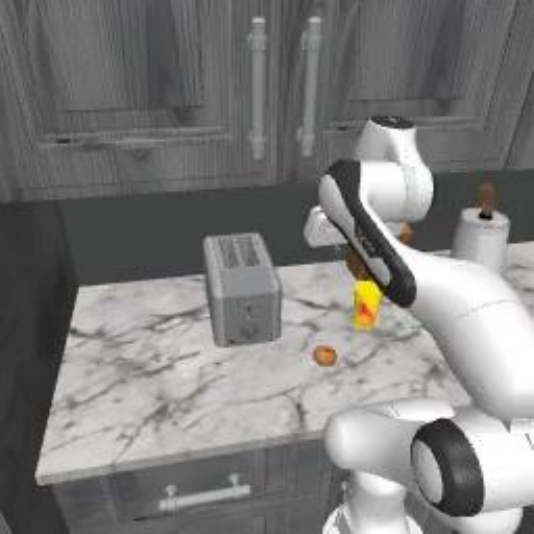}
    \caption{OpenCabinet}
\end{subfigure}
\begin{subfigure}{0.325\linewidth}
    \centering
    \includegraphics[width=0.7\linewidth]{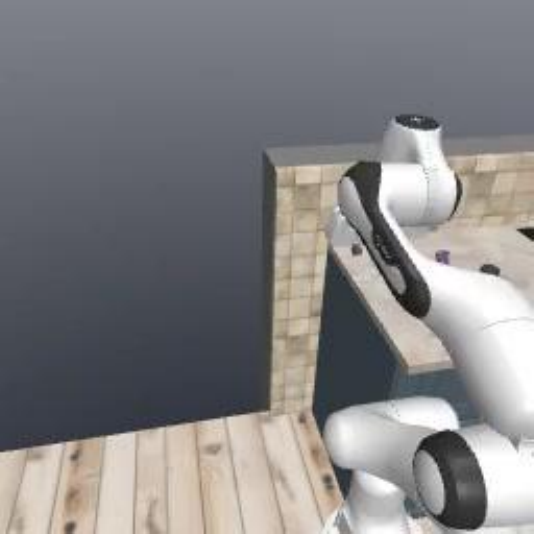}
    \caption{OpenDrawer}
\end{subfigure}

\vspace{0.35em}

\begin{subfigure}{0.325\linewidth}
    \centering
    \includegraphics[width=0.7\linewidth]{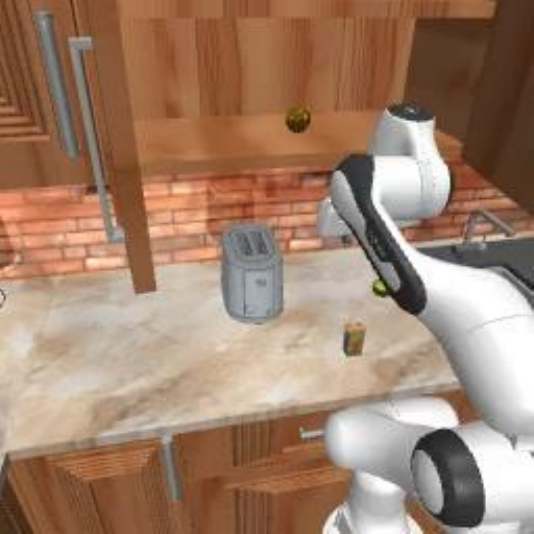}
    \caption{PickPlaceCounterToCabinet}
\end{subfigure}
\begin{subfigure}{0.325\linewidth}
    \centering
    \includegraphics[width=0.7\linewidth]{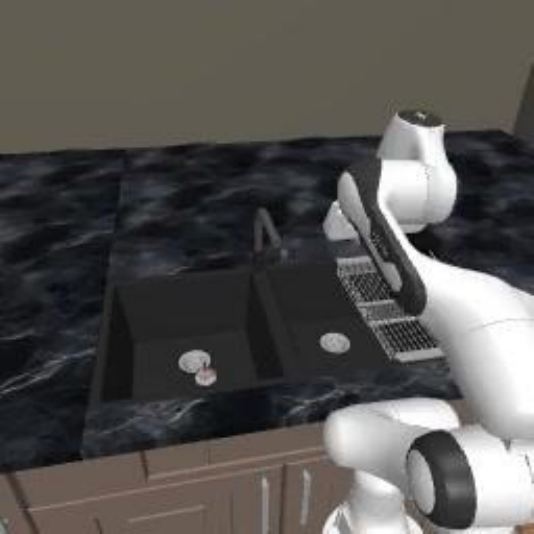}
    \caption{PickPlaceSinkToCounter}
\end{subfigure}
\begin{subfigure}{0.325\linewidth}
    \centering
    \includegraphics[width=0.7\linewidth]{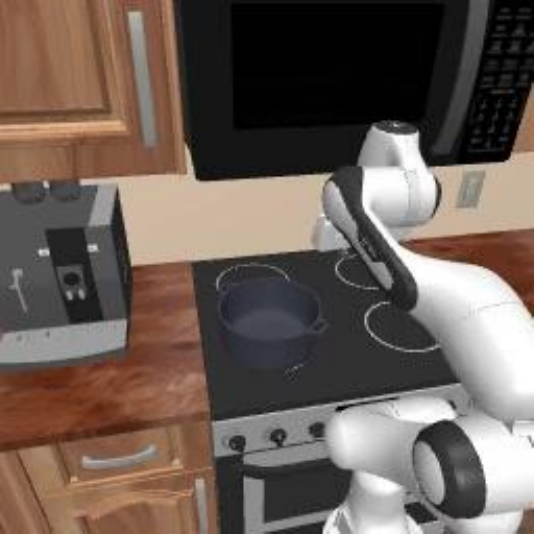}
    \caption{TurnOffStove}
\end{subfigure}

\vspace{0.35em}

\begin{subfigure}{0.325\linewidth}
    \centering
    \includegraphics[width=0.7\linewidth]{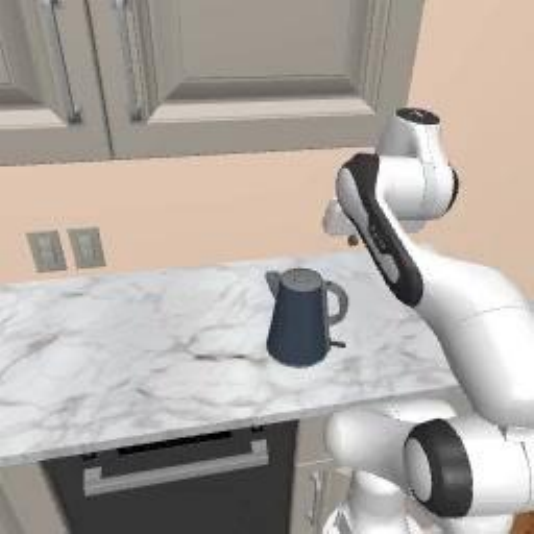}
    \caption{TurnOnElectricKettle}
\end{subfigure}
\begin{subfigure}{0.325\linewidth}
    \centering
    \includegraphics[width=0.7\linewidth]{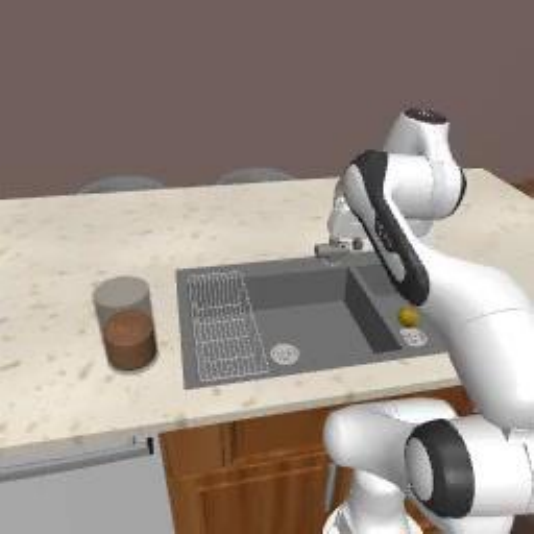}
    \caption{TurnOnSinkFaucet}
\end{subfigure}
\caption{
Visualization of the 11 RoboCasa tasks used in our evaluation. The tasks cover visually rich household scenes and robot manipulation, complementing the highly dynamic physics-control tasks in Kinetix.
}
\label{fig:app_robocasa_tasks}
\end{figure}

Figure~\ref{fig:app_robocasa_tasks} shows the 11 RoboCasa tasks used in our experiments. These tasks span a wide spectrum of household manipulation skills, ranging from simple reaching and grasping to multi-step sequential interaction with articulated objects. Although individual scenes are less dynamic than Kinetix, the long prediction horizons (chunk size 10, execution horizon 6) amplify the impact of chunk-boundary misalignment, making RoboCasa a complementary stress test for temporal consistency under asynchronous inference.

\clearpage

\begin{table}[t]
\centering
\caption{
Description of the 11 RoboCasa tasks used in our evaluation.
}
\label{tab:app_robocasa_task_descriptions}
\setlength{\tabcolsep}{4pt}
\renewcommand{\arraystretch}{1.05}
\resizebox{\linewidth}{!}{
\begin{tabular}{clp{0.62\linewidth}}
\toprule
\textbf{ID} & \textbf{Task} & \textbf{Manipulation property evaluated} \\
\midrule
1  & \texttt{CloseFridge}                & Pushing a fridge door shut; requires precise contact force and smooth deceleration at chunk boundaries to avoid bounce-back. \\
2  & \texttt{CloseToasterOvenDoor}       & Closing a small articulated door; constrained joint motion amplifies boundary discontinuity under delayed execution. \\
3  & \texttt{CoffeeSetupMug}             & Placing a mug under a coffee machine; multi-step alignment task where early chunk errors propagate to later stages. \\
4  & \texttt{NavigateKitchen}            & Mobile-base navigation in a cluttered kitchen; long-horizon trajectory where stale observations cause spatial drift. \\
5  & \texttt{OpenCabinet}                & Pulling open a cabinet door; requires stable chunk transitions during the grasp-then-pull sequence. \\
6  & \texttt{OpenDrawer}                 & Sliding a drawer open; linear constrained motion where jitter at chunk boundaries can jam the drawer track. \\
7  & \texttt{PickPlaceCounterToCabinet}  & Picking an object from the counter and placing it into a cabinet; two-phase task with a critical hand-off between chunks. \\
8  & \texttt{PickPlaceSinkToCounter}     & Transferring an object from the sink to the counter; requires grasp precision when the executed chunk lags behind the observed scene. \\
9  & \texttt{TurnOffStove}               & Rotating a stove knob to the off position; fine-grained rotational control sensitive to accumulated execution delay. \\
10 & \texttt{TurnOnElectricKettle}       & Pressing or toggling a kettle switch; small-displacement action where boundary jitter can miss the activation zone. \\
11 & \texttt{TurnOnSinkFaucet}           & Turning a faucet handle; requires smooth continuous rotation where chunk discontinuity causes incomplete activation. \\
\bottomrule
\end{tabular}
}
\end{table}

\subsection{Real-World Dynamic Manipulation Tasks}
\label{sec:app_real_world_tasks}

We conduct real-world experiments on two dynamic manipulation tasks. These tasks are designed to evaluate whether FreeBar can improve physical deployment when observations become stale during inference and when the robot must react continuously to moving or abruptly changing targets.

\paragraph{Conveyor-Belt Manipulation.}
The conveyor-belt manipulation task is performed on a dual-arm humanoid robot station. Objects are continuously transported by a moving conveyor belt, and the robot must detect, grasp, and sort the target objects into the designated container. This task is highly dynamic because the object pose changes throughout policy inference and action execution. If asynchronous execution directly executes the latest generated chunk without alignment, the chunk may correspond to an outdated object location, causing missed grasps, unstable contacts, or incorrect sorting. The task therefore evaluates whether the method can maintain chunk-boundary consistency while preserving the flexibility required to adapt to object motion.

\paragraph{Whack-a-Mole.}
The whack-a-mole task is performed with a Franka Emika robot. The robot must rapidly hit the currently active target within a fixed time window. The active target changes abruptly, so the policy must respond quickly to updated observations while avoiding discontinuous or oscillatory actions at chunk boundaries. We report the total number of successful hits within 30 seconds. This task directly tests real-time reactive control: excessive alignment may reduce reactivity, whereas insufficient alignment causes unstable motion and delayed responses.

\begin{figure}[t]
\centering
    \centering
    \includegraphics[width=\linewidth]{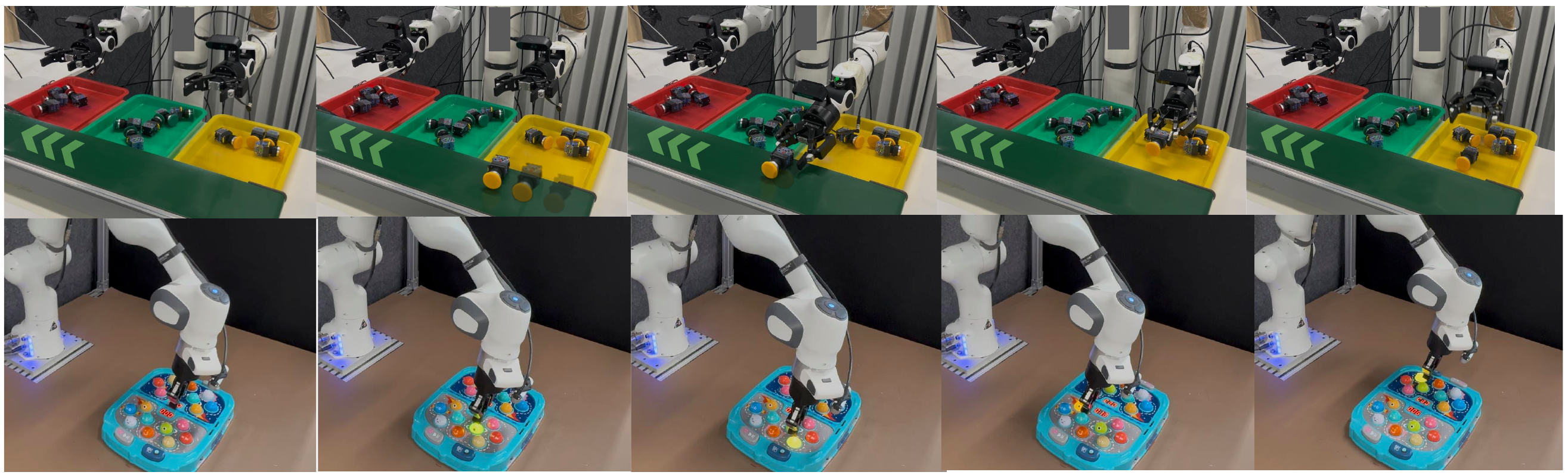}
    \caption{\textbf{Top}:Conveyor-Belt Manipulation \textbf{Bottom}:Whack-a-Mole
    Real-world dynamic manipulation tasks. Conveyor-Belt Manipulation evaluates dual-arm sorting of moving objects, while Whack-a-Mole evaluates single-arm reactive control under abrupt target changes. Both tasks expose severe state-action mismatch under asynchronous inference.
}
\label{fig:app_real_world_tasks}
\end{figure}

\begin{table}[t]
\centering
\caption{
Detailed description of the real-world tasks. Both tasks require real-time execution under non-stationary scene dynamics.
}
\label{tab:app_real_world_task_details}
\setlength{\tabcolsep}{4pt}
\renewcommand{\arraystretch}{1.08}
\resizebox{\linewidth}{!}{
\begin{tabular}{lp{0.22\linewidth}p{0.25\linewidth}p{0.30\linewidth}}
\toprule
\textbf{Task} & \textbf{Robot platform} & \textbf{Evaluation metric} & \textbf{Main dynamic challenge} \\
\midrule
Conveyor-Belt Manipulation 
& Dual-arm humanoid robot station 
& Sorting success rate over 30 rollouts 
& Target objects move during inference, so stale observations lead to grasp-location mismatch and sorting failure. \\
Whack-a-Mole 
& Franka Emika 7-DOF arm 
& Number of successful hits within 30 seconds 
& Active targets change abruptly, requiring rapid reaction while maintaining smooth chunk transitions. \\
\bottomrule
\end{tabular}
}
\end{table}

\subsection{Complete Hyperparameter Robustness Results}
\label{sec:app_hyperparam_robustness}

We evaluate robustness by sweeping the correction strength of RTC and FreeBar. For RTC, we sweep the VJP guidance strength $\beta_{\mathrm{RTC}}$. For FreeBar, we sweep the barrier projection strength $\alpha_{\mathrm{FreeBar}}$. Since the two parameters operate at different numerical scales and affect different correction mechanisms, we compare robustness at the method level using the mean and standard deviation across the tested configurations.

The complete results are reported in Tables~\ref{tab:app_hyperparam_robustness_sim}--\ref{tab:app_hyperparam_robustness_real_whack}. Across simulation and real-world tasks, FreeBar yields higher average performance and lower variance than RTC, indicating that closed-form barrier projection is less sensitive to task-specific tuning than VJP-based guidance.

\begin{table}[h!]
\centering
\caption{
Complete hyperparameter robustness results on RoboCasa simulation. Each entry reports the average task success rate over the evaluated RoboCasa tasks.
}
\label{tab:app_hyperparam_robustness_sim}
\setlength{\tabcolsep}{5pt}
\renewcommand{\arraystretch}{1.05}
{
\begin{tabular}{lcccccccc}
\toprule
\textbf{Method} & \textbf{0.1} & \textbf{0.5} & \textbf{1} & \textbf{5} & \textbf{10} & \textbf{30} & \textbf{50} & \textbf{Mean $\pm$ Std.} \\
\midrule
RTC     & 68.0 & 66.6 & -- & \textbf{70.0} & 69.6 & 68.6 & 52.8 & $65.93 \pm 6.55$ \\
FreeBar & 70.6 & -- & \textbf{71.4} & 70.4 & 70.6 & 67.6 & 67.2 & $69.63 \pm 1.77$ \\
\bottomrule
\end{tabular}
}
\end{table}

\begin{table}[h!]
\centering
\caption{
Complete hyperparameter robustness results on the real-world Conveyor-Belt Manipulation task with $\pi_0$. Each entry reports the sorting success rate (\%).
}
\label{tab:app_hyperparam_robustness_real_pi0}
\setlength{\tabcolsep}{5pt}
\renewcommand{\arraystretch}{1.05}
{
\begin{tabular}{lccccccccc}
\toprule
\textbf{Method} & \textbf{0.1} & \textbf{0.5} & \textbf{1} & \textbf{3} & \textbf{5} & \textbf{8} & \textbf{10} & \textbf{Mean $\pm$ Std.} \\
\midrule
RTC     & 53.33 & 56.67 & \textbf{67.67} & 53.33 & -- & 46.67 & -- & $55.53 \pm 7.70$ \\
FreeBar & 66.70 & -- & 73.30 & 76.67 & \textbf{83.33} & -- & 73.33 & $74.67 \pm 6.04$ \\
\bottomrule
\end{tabular}
}
\end{table}

\begin{table}[h!]
\centering
\caption{
Complete hyperparameter robustness results on the real-world Conveyor-Belt Manipulation task with $\pi_{0.5}$. Each entry reports the sorting success rate (\%).
}
\label{tab:app_hyperparam_robustness_real_pi05}
\setlength{\tabcolsep}{5pt}
\renewcommand{\arraystretch}{1.05}
{
\begin{tabular}{lccccccccc}
\toprule
\textbf{Method} & \textbf{0.1} & \textbf{0.5} & \textbf{1} & \textbf{3} & \textbf{5} & \textbf{8} & \textbf{10} & \textbf{Mean $\pm$ Std.} \\
\midrule
RTC     & 53.33 & 63.33 & \textbf{73.33} & 56.67 & -- & 50.00 & -- & $59.33 \pm 9.25$ \\
FreeBar & 70.00 & -- & 73.30 & \textbf{80.00} & \textbf{80.00} & -- & 76.67 & $75.99 \pm 4.35$ \\
\bottomrule
\end{tabular}
}
\end{table}

\begin{table}[h!]
\centering
\caption{
Complete hyperparameter robustness results on the real-world Whack-a-Mole task with $\pi_0$. Each entry reports the number of successful hits within 30 seconds.
}
\label{tab:app_hyperparam_robustness_real_whack}
\setlength{\tabcolsep}{5pt}
\renewcommand{\arraystretch}{1.05}
{
\begin{tabular}{lccccccccc}
\toprule
\textbf{Method} & \textbf{0.1} & \textbf{0.5} & \textbf{1} & \textbf{3} & \textbf{5} & \textbf{8} & \textbf{10} & \textbf{Mean $\pm$ Std.} \\
\midrule
RTC     & 5.01 & 6.95 & \textbf{10.08} & 8.95 & -- & 8.04 & -- & $7.80 \pm 1.92$ \\
FreeBar & 10.06 & -- & \textbf{12.03} & 9.95 & 11.02 & -- & 8.97 & $10.40 \pm 1.14$ \\
\bottomrule
\end{tabular}
}
\end{table}

\paragraph{Analysis.}
The robustness results show that FreeBar is less sensitive to the correction strength than RTC. RTC relies on VJP-based guidance, and its performance can vary substantially with the guidance coefficient. In contrast, FreeBar applies a closed-form action-space projection with progressive tightening and horizon-wise graded constraints. This design constrains only the temporally critical portion of the chunk while leaving distal actions flexible, which reduces the risk of over-correction. As a result, FreeBar achieves both higher mean performance and lower standard deviation across the tested hyperparameter settings.

\subsection{Discussion on Task Diversity and Dynamicity}
\label{sec:app_task_diversity_dynamicity}

The complete task suite is designed to evaluate asynchronous execution beyond a single controlled setting. Kinetix emphasizes fast-changing physics and contact-rich dynamics. RoboCasa emphasizes complex manipulation in visually rich household scenes. Conveyor-Belt Manipulation introduces continuous target motion in the real world, where delayed chunks easily become spatially misaligned with the object. Whack-a-Mole introduces abrupt target switching, requiring fast closed-loop reaction.

This coverage is important because asynchronous execution failure modes differ across domains. In dynamic simulation, failures often arise from delayed reactions to object motion. In household manipulation, failures often come from accumulated chunk-boundary discontinuities. In real-world conveyor sorting, the main issue is stale visual grounding for moving targets. In whack-a-mole, the main issue is reactive control under abrupt target changes.

\end{document}